\documentclass[runningheads]{llncs}
\usepackage{graphicx}

\usepackage{tikz}
\usepackage{comment}
\usepackage{amsmath,amssymb} 
\usepackage{color}

\usepackage{booktabs}
\usepackage{multirow}

\usepackage[pagebackref,breaklinks,colorlinks]{hyperref}
\usepackage[capitalize]{cleveref}
\crefname{section}{Sec.}{Secs.}
\Crefname{section}{Section}{Sections}
\Crefname{table}{Table}{Tables}
\crefname{table}{Tab.}{Tabs.}
\usepackage[accsupp]{axessibility}  

\begin{document}
\pagestyle{headings}
\mainmatter
\def\ECCVSubNumber{100}  

\title{Point Cloud Compression with Sibling Context and Surface Priors} 


\titlerunning{ }
%
\author{Zhili Chen \and
Zian Qian \and
Sukai Wang \and
Qifeng Chen}
\authorrunning{ }
%
\institute{The Hong Kong University of Science and Technology}
\maketitle

\begin{abstract}
We present a novel octree-based multi-level framework for large-scale point cloud compression, which can organize sparse and unstructured point clouds in a memory-efficient way. In this framework, we propose a new entropy model that explores the hierarchical dependency in an octree using the context of siblings' children, ancestors, and neighbors to encode the occupancy information of each non-leaf octree node into a bitstream. Moreover, we locally fit quadratic surfaces with a voxel-based geometry-aware module to provide geometric priors in entropy encoding. These strong priors empower our entropy framework to encode the octree into a more compact bitstream. In the decoding stage, we apply a two-step heuristic strategy to restore point clouds with better reconstruction quality. The quantitative evaluation shows that our method outperforms state-of-the-art baselines with a bitrate improvement of 11-16\% and 12-14\% on the KITTI Odometry and nuScenes datasets, respectively.
\end{abstract}

\section{Introduction}
LiDAR is undergoing rapid development and becoming popular in robots and self-driving vehicles. As the eyes of those machines, LiDAR sensors can generate point clouds that provide accurate 3D geometry in diverse environments. However, the large number of point clouds incur a heavy burden on storage and bandwidth usage. For example, a single Velodyne HDL-64E can generate more than 450 gigabytes of data over 30 billion points in just eight hours of driving. Therefore, developing effective algorithms for 3D point cloud compression is imperative.

Unlike image and video compression, point cloud compression is technically challenging due to the sparseness of orderless point clouds. In the early years, researchers utilize different data structures such as octrees~\cite{meagher1982geometric} and KD-trees~\cite{devillers2000geometric} to organize unstructured data and ignore the sparsity of point clouds. However, these methods did not reduce the information redundancy hidden in the data structure representations. Therefore, the potential of reducing the bitrate of a point cloud is still not well explored.

Recent works have shown that learning-based methods have great potential in reducing information redundancy. By encoding point cloud information into an embedded representation, a neural network is used to further reduce the bitrate by predicting a better probability distribution in entropy coding~\cite{DBLP:conf/cvpr/HuangWWLU20,que2021voxelcontext}. Their approaches can be summarized into three steps: (1) constructing an octree; (2) fusing different features such as ancestor information~\cite{DBLP:conf/cvpr/HuangWWLU20} or neighbor information~\cite{que2021voxelcontext} to construct a contextual feature map; (3) and training a shared weight entropy model to reduce the length of the bitstream. However, their approaches have three limitations. First, although previous approaches largely focus on the spatial dependencies such as ancestor information~\cite{DBLP:conf/cvpr/HuangWWLU20} or neighbor information~\cite{que2021voxelcontext}, they do not utilize prior knowledge of decoded siblings' children to reduce the information redundancy during the entropy coding. Second, they ignore local geometric information such as quadratic surface. Third, as illustrated in Fig.~\ref{fig:Heatmap}, since the distributions of occupancy symbols at different levels in an octree are very diverse, training a shared weight entropy model can overfit the octants at deeper levels and lead to low prediction accuracy on octants at shallower levels. 

To address these issues, we propose a novel octree-based multi-level framework. Our framework encodes the point cloud data into the non-leaf octants as eight-bit occupancy symbols. The entropy model encodes each occupancy symbol into a more compact bitstream through entropy coding by accurately estimating symbol occurrence probabilities. Compared to the previous methods, our entropy model is the first to utilize the siblings' children as strong priors when inferring the current octant's symbol probabilities. Our entropy model also incorporates the context of the current octant's neighbor and ancestor to explore the hierarchical contextual information of the octree fully. We also propose a quadratic surface fitting module to provide geometric priors, which is empirically proven to be beneficial in lowering the bitrate. In our compression framework, we train an independent entropy model for each octree level to capture resolution-specific context flow across different levels. At the decoding stage, we further improve the reconstructed point cloud quality by a two-step heuristic strategy that first predicts leaf octants' occupancies to retrieve the missing points by a specific level of entropy model and then apply a refinement module on the aggregated point cloud.

The contributions of this work can be summarized as follows.
\begin{itemize}
 \renewcommand{\labelitemi}{\textbullet}
 \item We build a novel octree-based multi-level compression framework for the point cloud. Our approach is the first one that incorporates sibling context for point cloud compression.
 \item We introduce surface priors into our entropy model, which effectively reduces octree-structured data's overall bitrate.
 \item In the decoding stage, we propose a two-step heuristic strategy by first predicting the missing points and then refining the aggregated point clouds to achieve a better reconstruction quality.
 \item Our proposed multi-level compression framework outperforms previous state-of-the-art methods on compression performance and reconstructs high-quality point clouds on two large-scale LiDAR point cloud datasets.
\end{itemize}

\section{Related Work}

\subsection{Traditional Point Cloud Compression}
Traditional point cloud methods usually utilize tree-based data structure to organize unstructured point cloud data, especially on KD-tree~\cite{devillers2000geometric,Draco} and Octree~\cite{garcia2018intra,graziosi2020overview,huang2008generic,jackins1980oct,kammerl2012real,schnabel2006octree,schwarz2018emerging}. Meagher et al.~\cite{meagher1982geometric} directly utilize octree to encode the point cloud without reducing the information redundancy. Huang et al.~\cite{huang2008generic} utilize the neighbourhood information for better entropy encoding. To reduce spatial redundancy, Kammerl et al.~\cite{kammerl2012real} use XOR to encode sibling octants. In recent years, MPEG also developed an octree-based standard point cloud compression method G-PCC~\cite{graziosi2020overview}. Moreover, Google utilizes KD-tree in its open-source point cloud compression software Draco~\cite{Draco}. Since all of the methods are hand-crafted, they cannot be implicitly optimized end-to-end. As such, the reduction of information redundancy is likely to be sub-optimal.

\subsection{Learning-based Point Cloud Compression}
Since point cloud data are in $n\times3$ unstructured floating points, it is almost impossible to directly apply convolutions to raw point cloud data. Thus, researchers proposed several kinds of methods to organize raw point clouds. Point-based methods~\cite{huang20193d,DBLP:journals/corr/abs-1905-03691} are usually built based on PointNet framework~\cite{Qi2017PointNetDH,qi2017pointnet}. Some point based methods~\cite{wiesmann2021deep} directly downsample point cloud to small set of points, then use deconvolution to rebuild the point cloud. However, point-based methods usually suffer from the sparsity of point cloud data and incur high memory costs in processing. Therefore, it cannot handle extremely large LiDAR point cloud data. Range-image-based point cloud compression methods~\cite{ahn2014large,nenci2014effective,sun2019novel,sun2020novel,tu2019point,DBLP:journals/corr/abs-1909-12037,DBLP:journals/corr/abs-2109-07717} first transfer point cloud into depth map or range image, then utilize state-of-the-art image compression methods such as ~\cite{DBLP:conf/iclr/TheisSCH17} for compression. However, such kinds of methods highly depend on the quality of image compression algorithms, and could not utilize spatial information. Voxel-based methods~\cite{biswas2020muscle,DBLP:conf/cvpr/HuangWWLU20,nguyen2021multiscale,que2021voxelcontext} ignore the sparsity of the point cloud data, and can utilize diverse geometric and spatial information. However, Huang et al.~\cite{DBLP:conf/cvpr/HuangWWLU20} only utilize the ancestor contexts, Que et al.~\cite{que2021voxelcontext} only use the neighbor contexts, and Fu et al.~\cite{OctAttention} ignore the local geometric information.

Unlike previous methods, our model is the first to reduce the inter-voxel redundancy by utilizing the context of decoded siblings' children octants. Moreover, local geometric information is also provided as the features for entropy model encoding. We also incorporate both ancestor and neighbor information into our entropy model to enrich the contextual information. 

\section{Methodology}
\begin{figure*}[t!]
\centering
\includegraphics[width=1.0\textwidth]{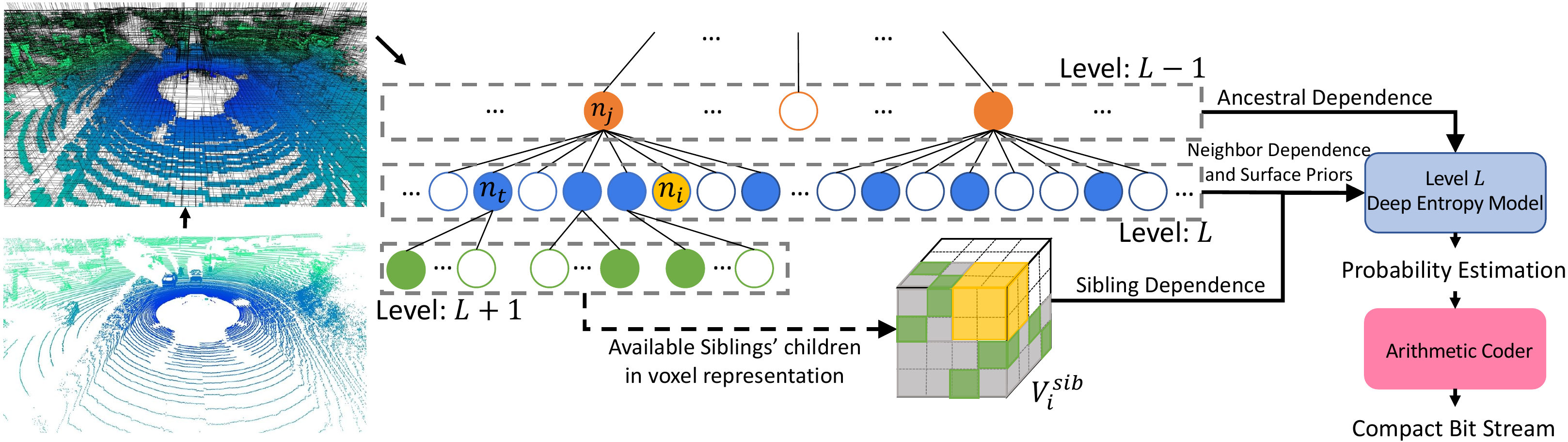}
\vspace{-1.5em}
\caption{The overview of our method. The input point cloud is first encoded into an octree. The ancestors, neighbors, and siblings' children octants are represented in orange, blue, and green colors. Our entropy model estimates occupancy probability distribution for each non-empty octant, conditioning on ancestral dependence, neighbor dependence, sibling dependence, and surface priors. Finally, the octree symbols are encoded into a more compact bitstream with arithmetic coding. The voxel block $V_i^{sib}$ is formed by decoded siblings' children. The green and the gray voxels in $V_i^{sib}$ represent the existent and absent siblings' children, respectively. The white voxels represent the unknown occupancies. After encoding the octant $n_i$ in yellow, its occupancy symbol will be filled in the yellow voxels of $V_i^{sib}$.}
\label{fig:model}
\end{figure*}

\subsection{Overview}
Our compression framework is shown in Fig.~\ref{fig:model}. Given an input point cloud, we first organize it using an octree with occupancy symbols stored at each non-leaf octant. A corresponding deep entropy model takes the hierarchical contexts as input for each octant to predict its occupancy symbol probabilities with 256 classes. The hierarchical contexts consist of local neighbors, ancestors, and siblings' children. The local neighbor context and sibling context are represented in binary voxel blocks. The ancestor information is an extracted feature map passed from the upper level. Moreover, we also incorporate our entropy model with surface priors by locally fitting quadratic surfaces among the local neighbors. Our entropy model can accurately predict symbol probabilities with these strong priors and further compress the symbols into a more compact bitstream by entropy coding.

In the decoding stage, we first reconstruct an octree with $L$ levels from the compressed bitstream. We then propose a two-step heuristic strategy to produce a point cloud with better quality. As illustrated in Fig.~\ref{fig:refine}, for each leaf octant, we first reuse our deep entropy model at level $L+1$ to retrieve its missing points by decoding the top-1 prediction of the occupancy symbol. Finally, we apply a separately trained refinement module~\cite{que2021voxelcontext} to further refine the aggregated points by adding the predicted offsets.

\subsection{Octree Construction}
Due to the sparseness of the orderless point cloud data, we have to first organize it in a well-structured representation before subsequent processing. Unlike other popular structures such as voxel-based methods, an octree is more memory efficient as it only partitions the non-empty 3D space and ignores the empty space. Moreover, the tree structure can provide hierarchical spatial information for a better reduction of information redundancy during compression. Therefore, we utilize octree to organize the point clouds.

An octree is built by recursively partitioning the input space into eight non-empty subspaces of the same size until the predefined depth is reached. The octant represents a bounding subspace and uses its center as the point coordinate. Each non-leaf occupied octant consists of an 8-bit occupancy symbol, and each bit represents the existence of a corresponding child (one for existence and zero for the absence). With such representation, a point cloud can be expressed as a serialized 8-bit occupancy symbols stream level-by-level and facilitate further lossless compression by an entropy encoder.

\begin{figure*}[t!]
\centering
\includegraphics[width=1.0\textwidth]{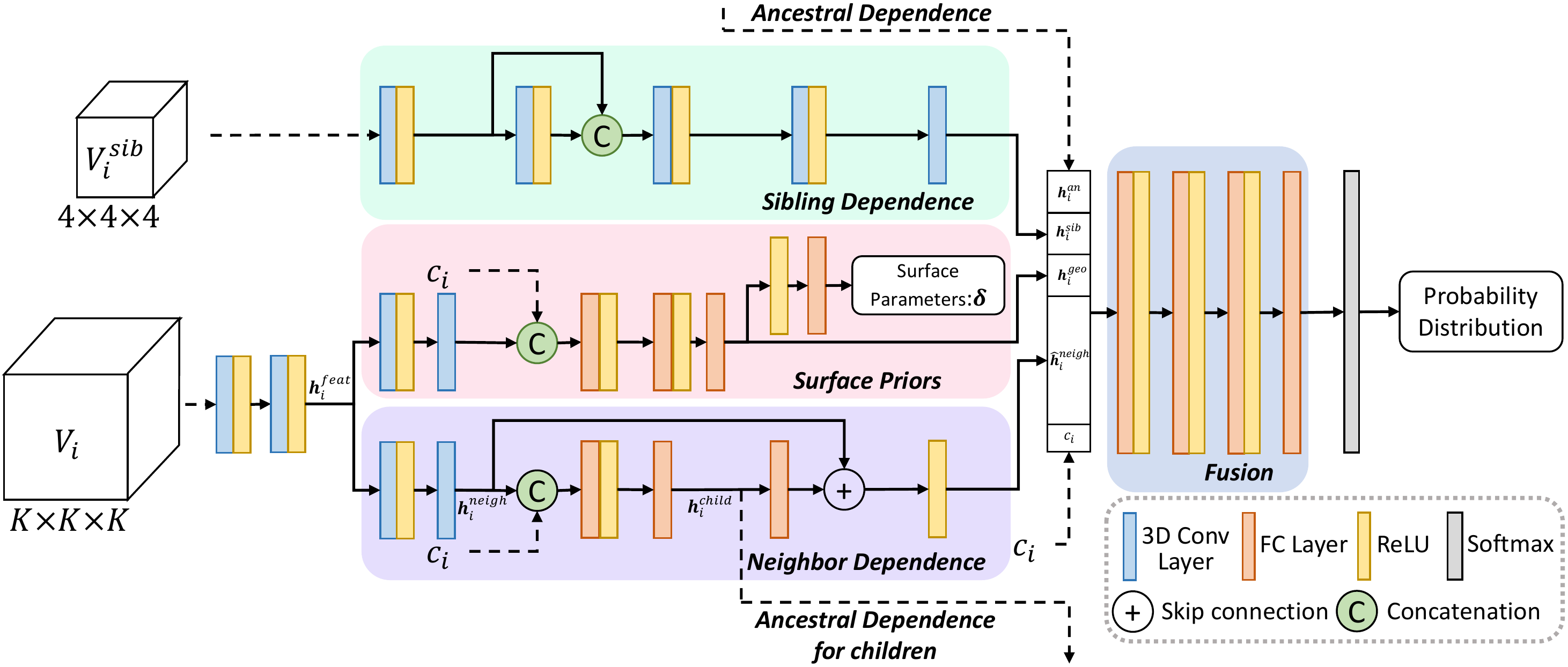}
\vspace{-1.5em}
\caption{The network architecture of our deep entropy model. The dash lines indicate the contextual input to our entropy model. The sibling and neighbor contexts are present as $V_i^{sib}$ and $V_i$ in voxel representations, $c_i$ is the octant information (located tree level and corresponding coordinates) and $h_i^{an}$ is the ancestor context. Note that the dimension $K$ of the neighbor context $V_{i}$ is empirically set to 9 in experiments.}
\label{fig:structure}
\end{figure*}

\subsection{Hierarchical Context Entropy Model}
Given an occupancy symbol stream $\boldsymbol{\mathrm{s}}=[s_1,s_2,...,s_n]$ with the number of non-leaf octants as its length $n$, the goal of our entropy model is to minimize the bitstream length. According to information theory, this goal can be reached by minimizing cross-entropy loss $\mathop{\mathbb{E}}_{s\sim p}[-\log q(\boldsymbol{\mathrm{s}})]$, where $q(s)$ is the estimated probability distribution of occupancy symbol $s$ and $p$ is the actual distribution.

The accuracy of the probability distribution $q(s)$ determines the effectiveness of the entropy model. For example, if the actual probability distribution of $s$ is known, no bit is needed to encode the whole point cloud. However, formulating the actual $q(s)$
could be very difficult because it has many complex dependencies such as ancestor or neighbor context. A good formulation of $q(s)$ can provide strong priors for the entropy model and therefore reduce the information redundancy in a bitstream. As illustrated in Fig.~\ref{fig:model}, our entropy model utilizes hierarchical dependencies from coarse to fine with ancestors, neighbors, and decoded siblings' children. Moreover, we utilize surface information as geometric priors while estimating $q(s)$. Therefore, the formulation of $q(s)$ in our entropy model can be defined as 
\begin{equation}
q(\boldsymbol{\mathrm{s}})=\prod_i{q(s_i\mid h_{i}^{an}, h_{i}^{neigh}, h_{i}^{sib}, h_{i}^{geo}, c_i;\mathrm{\theta})},
\end{equation}
where $ h_{i}^{an}$ is the ancestral dependence, $ h_{i}^{neigh}$ is the neighbor dependence, $ h_{i}^{sib}$ is the siblings' children dependence, $ h_{i}^{geo}$ is the surface priors, $c_i$ is the octant information (located tree level and corresponding coordinates), and $\theta$ is the parameters of our entropy model.

\subsubsection{Neighbor Dependence}\label{neig_Dep}\hfill\\
The octree at depth $l$ ($l\in[1,L]$) can be considered as a discretization of the 3D space at the resolution of $2^{l}\cdot2^{l}\cdot2^{l}$. Inspired by~\cite{que2021voxelcontext}, we construct the neighbor context of an octant $n_i$ by locally forming a $K\times K\times K$ binary voxel block $V_i$ centered at $n_i$ and $K$ is empirically set to 9 in experiments. Each binary value indicates the existence of its corresponding neighbors. The neighbor voxel $V_{i}$ is transformed by a feature extraction function $f_{neigh}$:

\begin{equation}
h^{neigh}_{i}=f_{neigh}(V_i),
\end{equation}
where $h^{neigh}_i$ is the output feature vector that represents the neighbor contextual information. The feature vector $h^{neigh}_i$ is provided as part of the prior knowledge for inferring the $n_i$'s children distribution.

\subsubsection{Ancestral Dependence}\hfill\\
The ancestral information brings coarser geometric information from a shallower level to the current octant~\cite{DBLP:conf/cvpr/HuangWWLU20}, which can enlarge the entropy model's receptive field. As illustrated in Fig.~\ref{fig:structure}, since the entropy model is trained level-by-level, the neighbor context features $h_{i}^{neigh}$ of current octant $n_{i}$ can be further passed to its children as ancestral features.

To avoid ancestral features overwhelming the feature space of its children octants, we first concatenate the $h_{i}^{neigh}$ with the octant information $c_i$ and then apply an MLP denoted as $\varphi$ to extract more condensed features $h^{child}_{i}$. The processed features $h^{child}_{i}$ will then be passed to its children as their ancestral dependence. A following linear projection layer $\gamma$ is applied on $h^{child}_{i}$ to scale back the features to the original dimension. Then the feature is added with $h_{i}^{neigh}$ through skip connection to recover the original neighbor feature $\hat{h}^{neigh}_{i}$ of the current octant:

\begin{equation}
h^{child}_{i}=\varphi({\rm{Concat}}(h^{neigh}_{i},c_i)),\hat{h}^{neigh}_{i}=\gamma(h^{child}_{i})+h^{neigh}_i.
\end{equation}

Note that the ancestral features received from the upper level for the current octant is denoted as $h^{an}_{i}$. For the root octant, we initialize its ancestral features with zero values.

\subsubsection{Sibling Context}\hfill\\
Sibiling octants are adjacent in the original 3D space. The decoded children of siblings are represented in a finer voxel size and are even closer to the encoding/decoding octant. Similar to successive pixels in 2D images and videos, siblings' children are strongly correlated with the children of the current octant because the geometric structure of these siblings' children and the current octant's children can be quite similar. Therefore, conditioning on siblings' children can reduce the information redundancy while predicting the probability distribution of the current octant.

As illustrated in Fig.~\ref{fig:model}, an occupied octant $n_j$ is located at the octree level $L-1$ and represents a space in the 3D world. Until the level of $L+1$, this space is partitioned into $4\times4\times4$ subspaces. We represent it as a binary voxel  $V_i^{sib}$. Since our entropy model encodes the occupancy symbols of $n_j$’s children sequentially, the previous siblings’ occupancy symbols are available while predicting the probability of the current octant $n_i$. The available occupancy symbols are filled in $V_i^{sib}$. Note that the first occupied child octant of $n_t$ has no available sibling context, and it will take $V_i^{sib}$ with zero values as input. We parameterize another feature transformation function, $f_{sib}$, as a 3D convolution network to exploit the sibling context with $V_i^{sib}$ as input. We formulate it as
\begin{equation}
h^{sib}_{i}=f_{sib}(V^{sib}_i).
\end{equation}
The $h^{sib}_{i}$ is the output feature vector that represents the sibling contextual information and will be further provided as a part of prior knowledge for entropy coding.

\subsubsection{Geometric Priors}\hfill\\ 
Since LiDAR is a time-of-flight device to estimate the round-trip time of the darting laser beams reflecting from a scene, most of the sampling LiDAR points are densely distributed at large surfaces, such as roads and vehicles. If an octant crosses the boundary of a surface, its children octants are more likely to be occupied. Therefore, geometric information such as the surface can be a strong prior for reducing the data redundancies. 

The neighbor context of $n_i$ is defined as a voxel representation $V_i$ in Section~\ref{neig_Dep}. We first transform the neighbor context $V_i$ into a point representation $\left\{(x_i,y_i,z_i)\right\}^N_{i=1}$ with the coordinate of $n_i$ as the origin. Then we train a module to locally fit a quadratic surface by minimizing the vertical distance from each point to the surface:
\begin{equation}
\mathcal{L}_{sf}=\|\boldsymbol{\mathrm{z}}-\boldsymbol{\mathrm{\delta}}\left[\boldsymbol{\mathrm{x}}^2,\boldsymbol{\mathrm{y}}^2,\boldsymbol{\mathrm{x}}\boldsymbol{\mathrm{y}},\boldsymbol{\mathrm{x}},\boldsymbol{\mathrm{y}},\boldsymbol{\mathrm{1}}\right]^T\|^2_2,
\end{equation}
where $\boldsymbol{\mathrm{\delta}}\in\mathbb{R}^6$ are six parameters of the quadratic surface.
As illustrated in Fig.~\ref{fig:structure}, the module that learns surface priors will take the low level features $h_{i}^{feat}$ from the second layer of $f_{neigh}$ as input. The surface estimation module $f_{geo}$ extract the geometric features from $h_{i}^{feat}$, followed by a MLP to estimate the parameters $\boldsymbol{\mathrm{\delta}}$ for the quadratic surface. The transformation can be defined as

\begin{eqnarray}
h_i^{geo}&=&f_{geo}(h^{feat}_{i}),\\
\boldsymbol{\mathrm{\delta}}&=&\boldsymbol{{\rm MLP}}(h_i^{geo}),
\end{eqnarray}
where $h_i^{geo}$ represents the geometric priors provided to the entropy model to infer the probability distribution.

\subsubsection{Entropy Model Header}\hfill\\ Eventually, the neighbor feature $h^{neigh}_i$, the ancestral feature $h^{an}_i$, siblings' children feature $h^{sib}_i$, the geometric feature $h_i^{geo}$, and the current octant's information $c_i$ are aggregated to a four-layer MLP followed by a softmax to predict a 256 channels probability $q(s_i)$ for the 8-bit occupancy symbol. Then we minimize the cross-entropy loss on each octant:
\begin{equation}
\mathcal{L}_{CE}= -\sum_i p(s_i)\log q(s_i),
\end{equation}
where $p(s_{i})$ is the ground-truth probability distribution of the occupancy symbol $s_{i}$.

\subsection{Multi-level Learning Framework}
\begin{figure}[t!]
\centering
\includegraphics[width=0.4\textwidth]{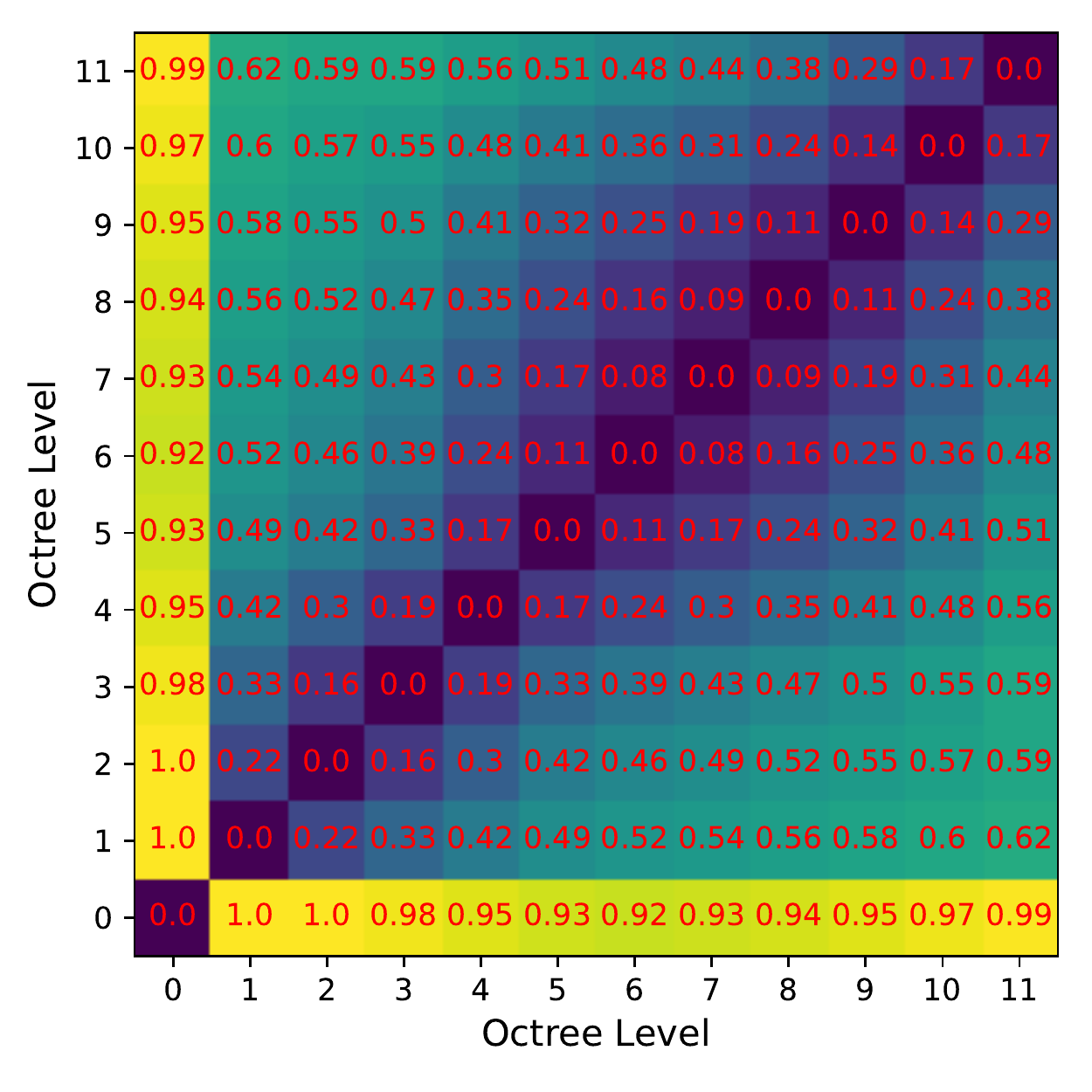}
\vspace{-1.5em}
\caption{The heat map of the Jensen-Shannon (JS) divergence of the occupancy symbol distributions among different levels of an octree. The distributions are computed from our training set on KITTI~\cite{Geiger2012CVPR}. JS divergence is to measure the difference between two probability distributions. The greater JS divergence means the larger difference between the two distributions.}
\label{fig:Heatmap}
\end{figure}
From our statistics on 12-level octrees constructed from the KITTI Odometry~\cite{Geiger2012CVPR} dataset, the last level of the non-leaf octants accounts for more than 50\% of the total number of all the non-leaf octants. We calculate the occupancy symbols distribution by levels and compute the Jensen–Shannon (JS) divergence among them. As shown in Fig.~\ref{fig:Heatmap}, the greater JS divergence means the larger difference
between the two distributions.

Based on these two observations, we separately train entropy models for each level instead of training a single shared-weight entropy model on the entire octree. This multi-level learning framework facilitates our entropy model to better reduce each octant's spatial redundancies in different spatial resolution.

We use the same objective function for the entropy models at different levels:
\begin{equation}
\mathcal{L}= \mathcal{L}_{CE}+\lambda\mathcal{L}_{sf},
\end{equation}
where $\lambda$ is the weight of the loss and empirically set to 0.2 in experiments.

\begin{figure}[t!]
\centering
\includegraphics[width=1.0\textwidth]{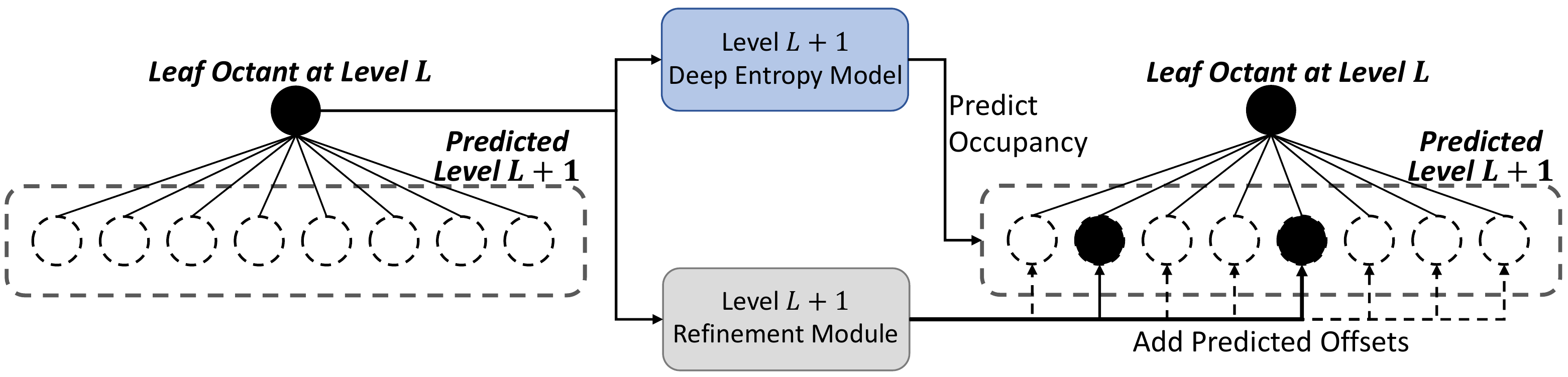}
\vspace{-2.0em}
\caption{The illustration of our two-step refinement strategy. The leaf octant of the decoded octree locates at the depth level L, and its occupancies are unknown. We first apply our entropy model on the octant to predict the occupancy symbols and calculate their coordinates. Then the coordinates are added with the corresponding offsets predicted from the refinement module.}
\label{fig:refine}
\end{figure}

\subsection{Two-step Reconstruction Strategy}
The quantization error of our point cloud compression framework comes from the octree construction and depends on the octree level $L$. Each leaf octant at level $L$ represents a subspace in the large 3D space. Because of the quantization, we can only recover a single point from each leaf octant by taking the center coordinate of its corresponding subspace. VoxelContext~\cite{que2021voxelcontext} introduces a refinement module to reduce the quantization error by adding predicted offsets to these coordinates in the decoding stage. In contrast, we approach this problem by first retrieving missing points with the help of our trained entropy model at level $L+1$ and then further refine the aggregated points. By this strategy, we can reconstruct point clouds with the precision near to level $L+1$ while keeping the bitrate at level $L$.

As illustrated in Fig.~\ref{fig:refine}, considering a reconstructed octree with $L$ levels, we denote the set of leaf octants as $\mathcal{N}_L$. We first apply our entropy model at level $L+1$ to estimate the occupancy symbol probabilities for each leaf octant in $\mathcal{N}_L$. We take the predicted occupancy symbols $\mathcal{S}_{L+1}$ with top-1 accuracy to form a new set of octants $\mathcal{N}_{L+1}$ located at the predicted level $L+1$. We denote the represented 3D coordinate of $\mathcal{N}_{L+1}$ as $\boldsymbol{\mathrm{x}}^p\in\mathbb{R}^{n\times3}$. In the second step, we apply the refinement module at level $L+1$ ~\cite{que2021voxelcontext} that takes neighbor context $V_i$ and octant information $c_i$ of leaf as an input to predict offsets for the coordinates. Our refinement module is predefined to output offsets for every children octant of $\mathcal{N}_{L}$, where the offsets are denoted as $\boldsymbol{\mathrm{x}}^o$. Then we obtain the final output of the coordinate by
\begin{equation}
\boldsymbol{\mathrm{x}}^r=\boldsymbol{\mathrm{x}}^p+\boldsymbol{\mathrm{I}}(\mathcal{S}_{L+1}, \boldsymbol{\mathrm{x}}^o),
\end{equation}
where $\boldsymbol{\mathrm{I}}(a,b)$ denote the operation of indexing $b$ according to $a$. To be specifically, $\boldsymbol{\mathrm{I}}(\mathcal{S}_{L+1}, \boldsymbol{\mathrm{x}}^o)$ means that for those octants whose occupancy codes are zeros, we ignore their offsets. We only add the offsets to the octants where their occupancy code are ones.
$\boldsymbol{\mathrm{x}}^r\in\mathbb{R}^{n\times3}$ is the resulting refined 3D coordinates.

To obtain the ground-truth coordinate of $N_{L+1}$ during the training of entropy model, We build octrees with $L+1$ levels and calculate the coordinates $\boldsymbol{\mathrm{x}}^{gt}$ from $\mathcal{N}_{L+1}$ as the ground truth. Note that the sibling context at level $L+1$ is not available during the encoding and decoding time, and we set $V_i^{sib}$ as zeros voxel block to be the input of our entropy model. We train the refinement module $\mathcal{R}$ by minimizing the Chamfer Distance~\cite{fan2017point} of the $\boldsymbol{\mathrm{x}}^{gt}$ and $\boldsymbol{\mathrm{x}}^r$:

\begin{equation}
\mathcal{L}_{CD}=\max\left\{\text{CD}(\boldsymbol{\mathrm{x}}^{r},\boldsymbol{\mathrm{x}}^{gt}),\text{CD}(\boldsymbol{\mathrm{x}}^{gt},\boldsymbol{\mathrm{x}}^{r})\right\},
\end{equation}
where $\mathcal{L}_{CD}$ is the objective function for the refinement module.

\section{Experiments}
\subsection{Experimental Setup}
\textbf{Dataset.}
We evaluate the compression performance and reconstruction quality on two real-world datasets with different densities: KITTI Odometry~\cite{Geiger2012CVPR} and nuScenes~\cite{caesar2020nuscenes}, which are collected by a Velodyne HDL-64 sensor and a Velodyne HDL-32 sensor. We also use the KITTI detection dataset~\cite{Geiger2012CVPR} for evaluating the 3D object detection as a downstream task using the reconstructed point cloud.

To ensure the diversity of the training and testing scene, we randomly sample 6000 frames from sequence 00 to 10 in the KITTI Odometry dataset for training and 550 frames from sequence 11 to 21 for testing. For nuScenes, we randomly sample 1200 
frames from each of the first five data batches (total 6000) for training, and randomly sampled 100 frames from each of the last five data batches (total 500) for testing. Note that we compare our compression performance and reconstruction quality with the original raw point cloud without any pre-processing in the following sections.

\textbf{Baselines.}
To evaluate the performance of our framework, we compare our method with three state-of-the-art baselines: hand-drafted Octree-based point cloud compression method MPEG G-PCC~\cite{schwarz2018emerging}, KD-tree based method Google Draco~\cite{Draco}, and also a learning-based method VoxelContext~\cite{que2021voxelcontext}. VoxelContext has been faithfully self-implemented as the official code has not been released.

\textbf{Implementation details.} We construct the octree with a maximum level of 13 on both KITTI Odometry and nuScenes datasets. Note that we construct level 13 of the octree only for generating the ground-truth labels for our training. In our experiment section, we evaluate the performance of our framework by truncating octree levels ranging from 9 to 12 on both datasets to vary the compression bitrates. The corresponding spatial quantization error ranges from 14.94 cm to 1.87 cm for KITTI Odometry and 18.29 cm to 2.29 cm for nuScenes.

We implement our framework in PyTorch and experiments on a machine with two NVIDIA 3090 GPUs. The total number of parameters in our deep entropy model is 1.77M. We use Adam optimizer~\cite{DBLP:journals/corr/KingmaB14} with a learning rate of 1e-4 to train our whole framework. The training epoch for the multi-level entropy model and the refinement module are 20 and 2, respectively. To be memory efficient, we do not backpropagate the gradients of an entropy model from the lower to the upper level during the training of multi-level entropy models. We freeze the weights of the entropy models while training the refinement module.

\subsection{Evaluation Metrics}
We evaluate our framework from two aspects, compression performance, and reconstruction quality. Bits per point (BPP) is the most commonly used metric for evaluating the compression performance. Since we only consider the geometric compression of a point cloud in this section, the size of the original point cloud data is calculated by $96 \times N$, where $N$ is the number of points in the raw point cloud, and 96 is the size of the coordinates: $x$, $y$ and $z$, where each coordinate is represented in a 32-bit floating-point. BPP is defined as $BPP=\lvert bit\rvert/N$, where the $\lvert bit\rvert$ is the bitstream length.

We utilize four metrics to evaluate the reconstruction quality of the point cloud: point-to-point and point-to-plane PSNR ~\cite{mekuria2017performance,tian2017geometric}, F1 score, and maximum Chamfer Distance~\cite{fan2017point,tian2017geometric}. Same as the evaluation metrics defined in ~\cite{biswas2020muscle}, for original input point cloud $P$ and reconstruction point cloud $\hat{P}$, we use $\text{PSNR} = 10\log_{10}\frac{3r^{2}}{\max\{\text{MSE}(P, \hat{P}), \text{MSE}(\hat{P}, P)\}}$, $\text{F1} = \frac{TP}{TP+FP+FN}$, $\text{CD}_{max} = \max\{\text{CD}(P, \hat{P}), $ $\text{CD}(\hat{P}, P)\}$, where the peak constant value of $r$ is 59.70 m, and the distance threshold of the F1 score is 2 cm.
\subsection{Results}

\begin{figure*}[t]
\centering
\begin{tabular}{@{}c@{\hspace{0.2mm}}c@{\hspace{0.2mm}}c@{\hspace{0.2mm}}c@{\hspace{0.2mm}}c@{\hspace{0.2mm}}c@{\hspace{0.2mm}}c@{}}
\hspace{-2mm}
\rotatebox{90}{\scriptsize \hspace{1mm} KITTI Odometry}&
\includegraphics[width=0.25\linewidth]{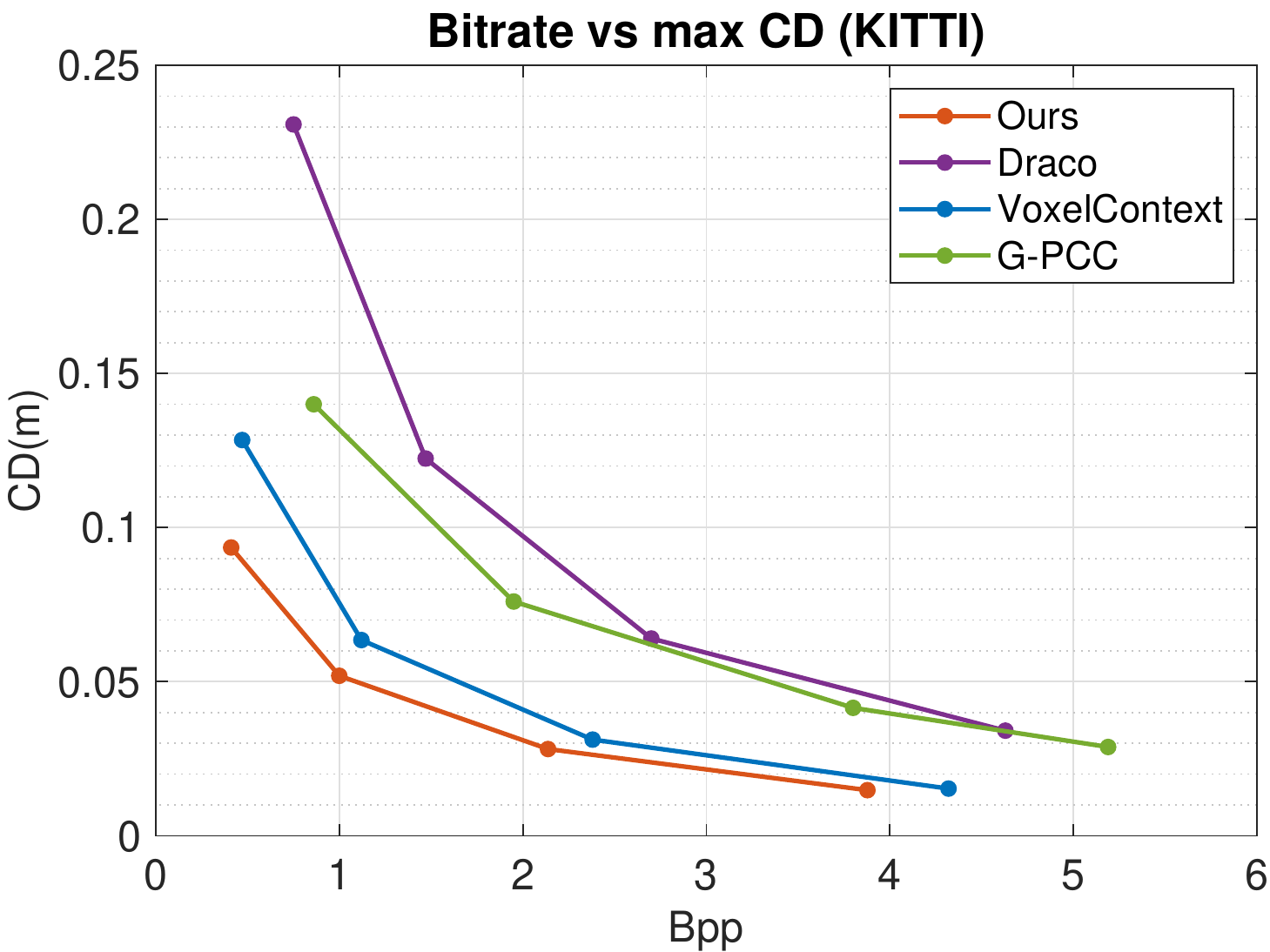}&
\includegraphics[width=0.25\linewidth]{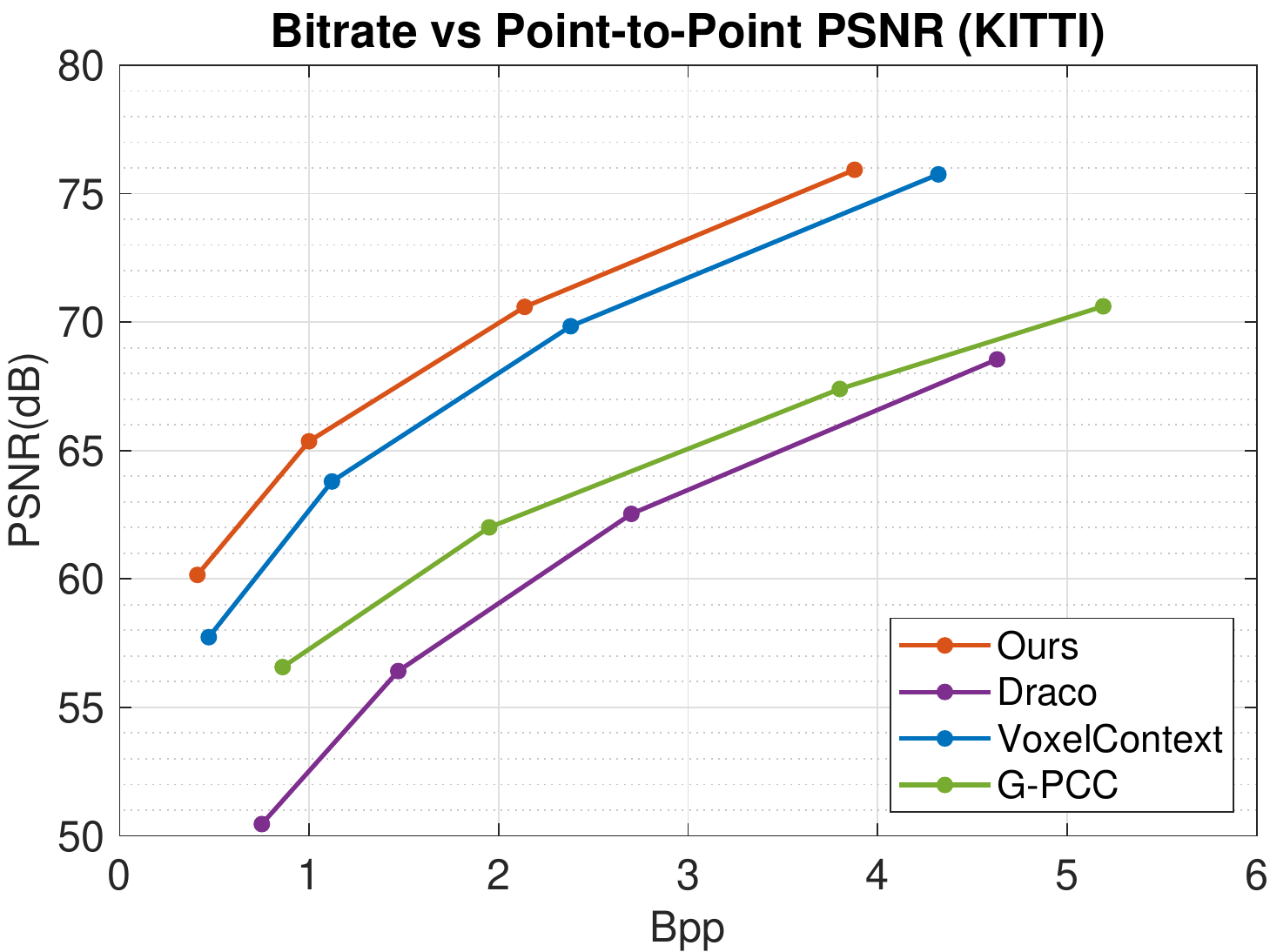}&
\includegraphics[width=0.25\linewidth]{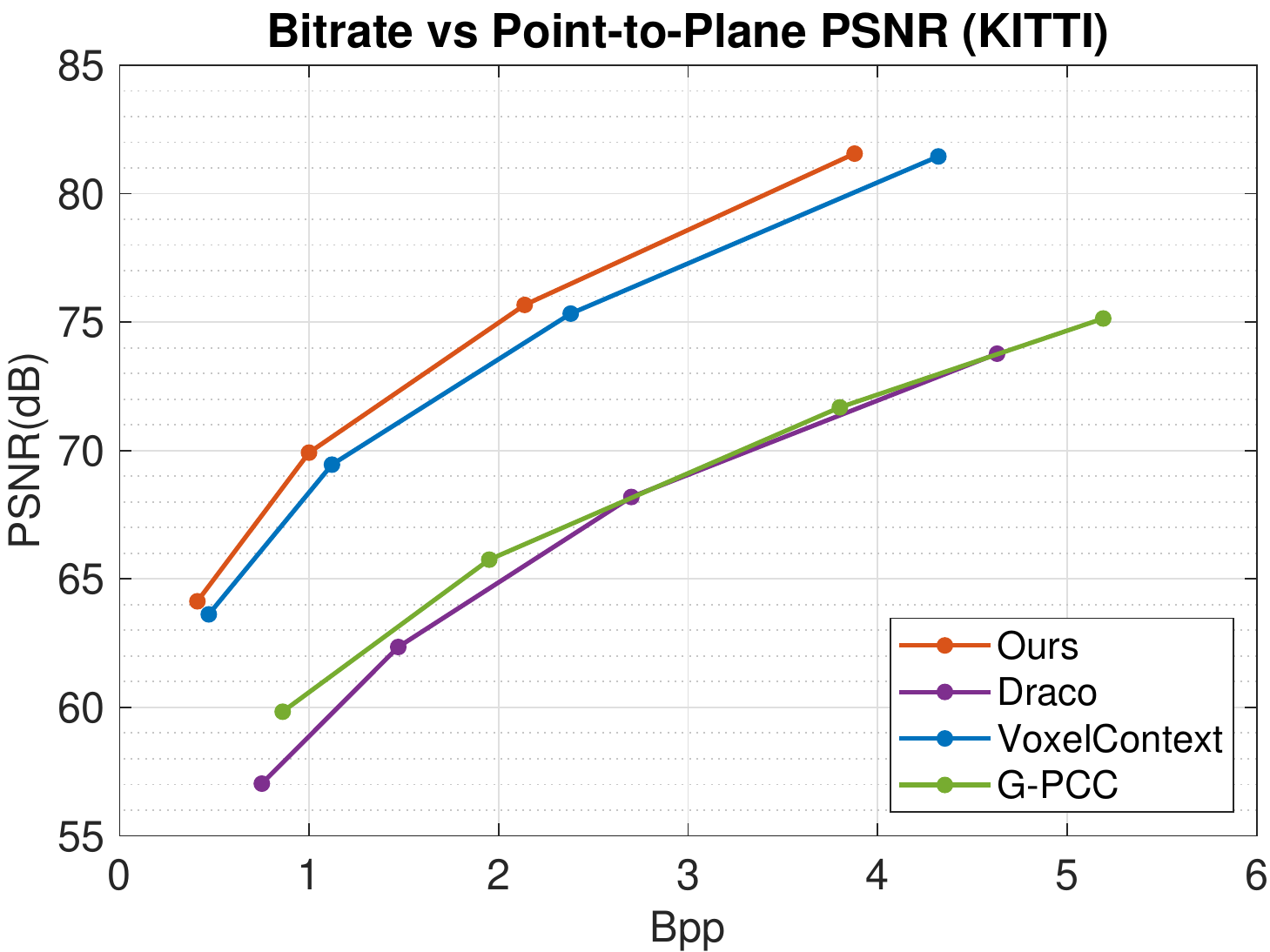}&
\includegraphics[width=0.25\linewidth]{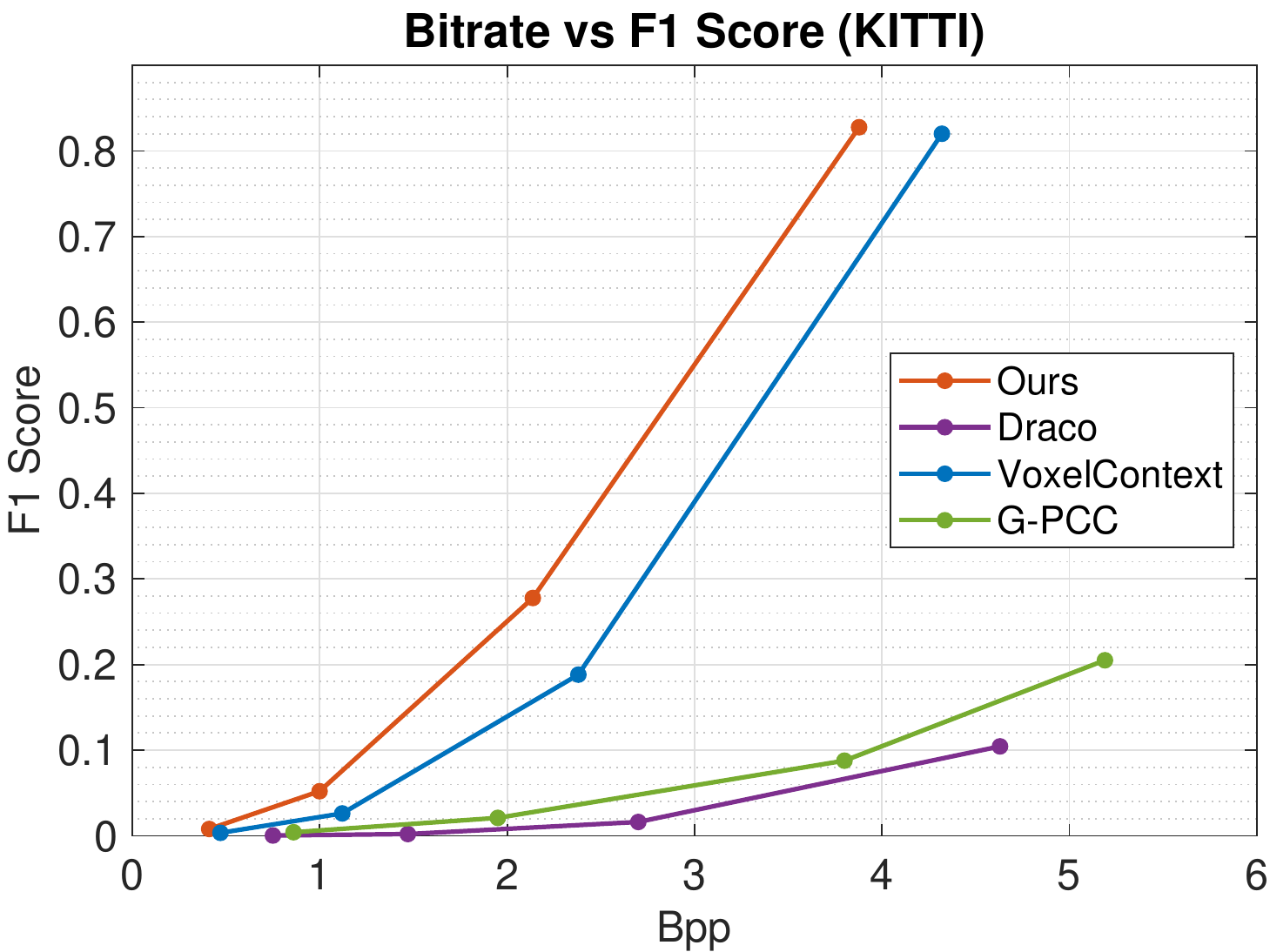}&\\
\hspace{-2mm}
\rotatebox{90}{\scriptsize \hspace{6mm} nuScenes}&
\includegraphics[width=0.25\linewidth]{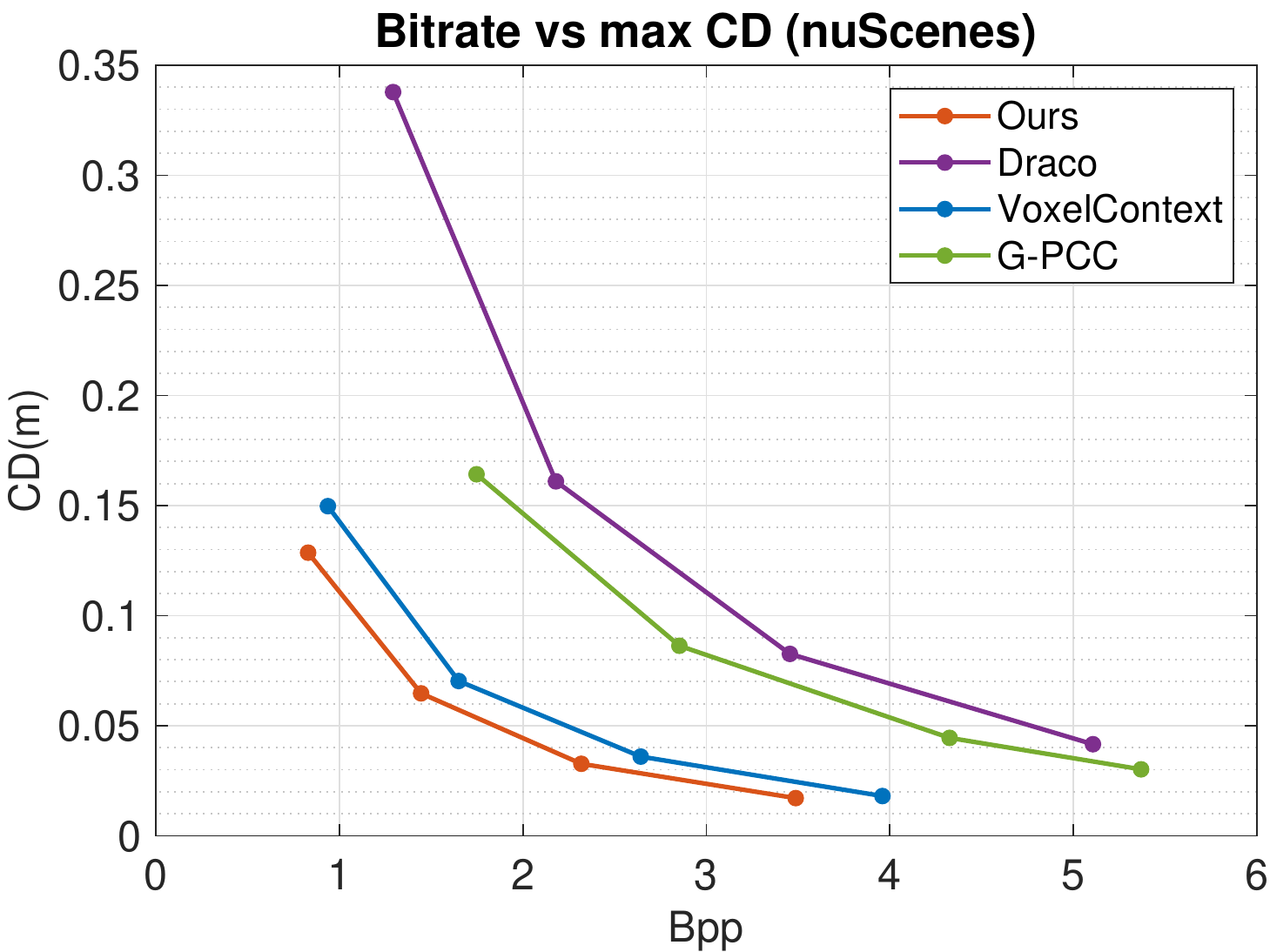}&
\includegraphics[width=0.25\linewidth]{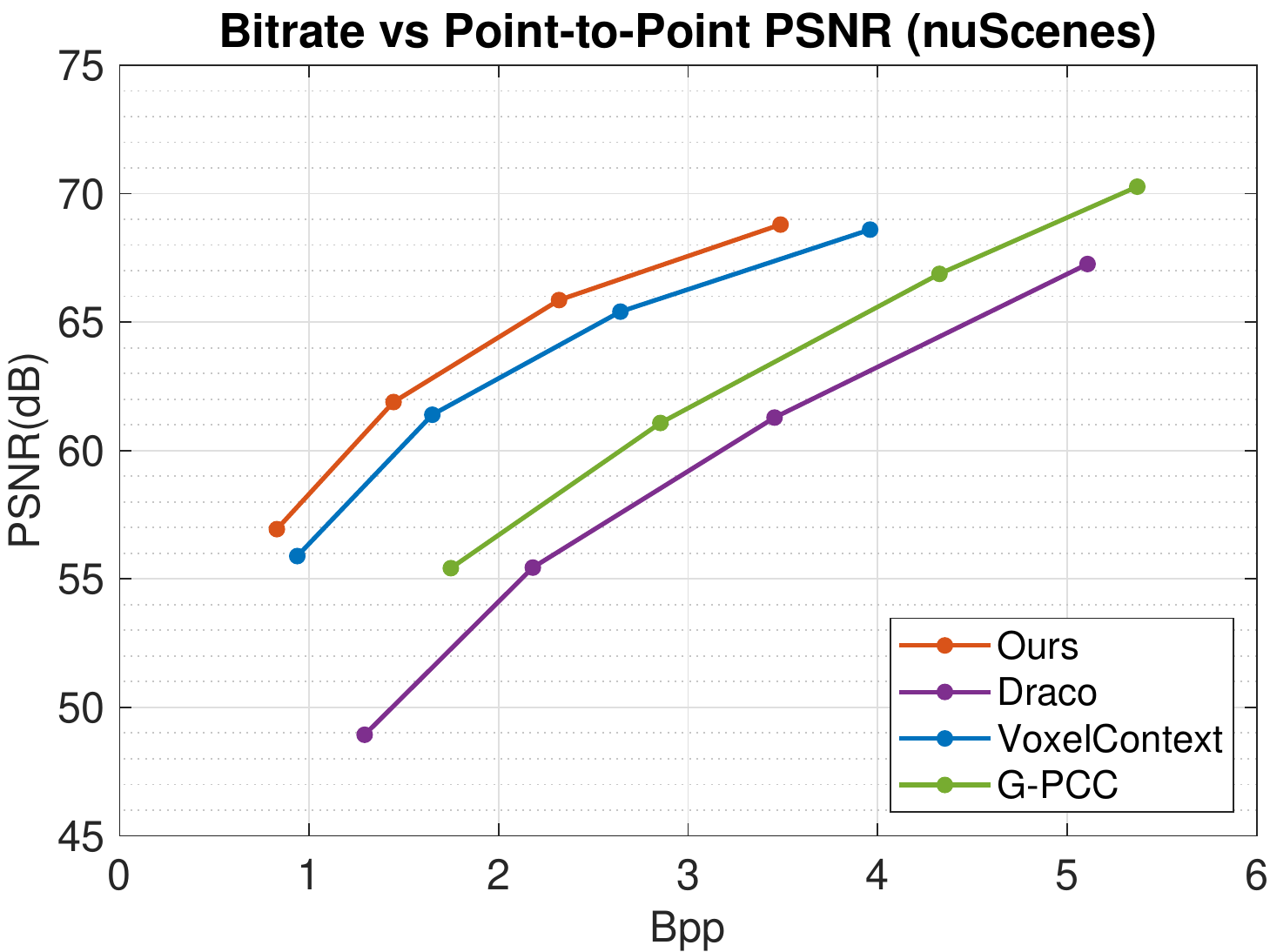}&
\includegraphics[width=0.25\linewidth]{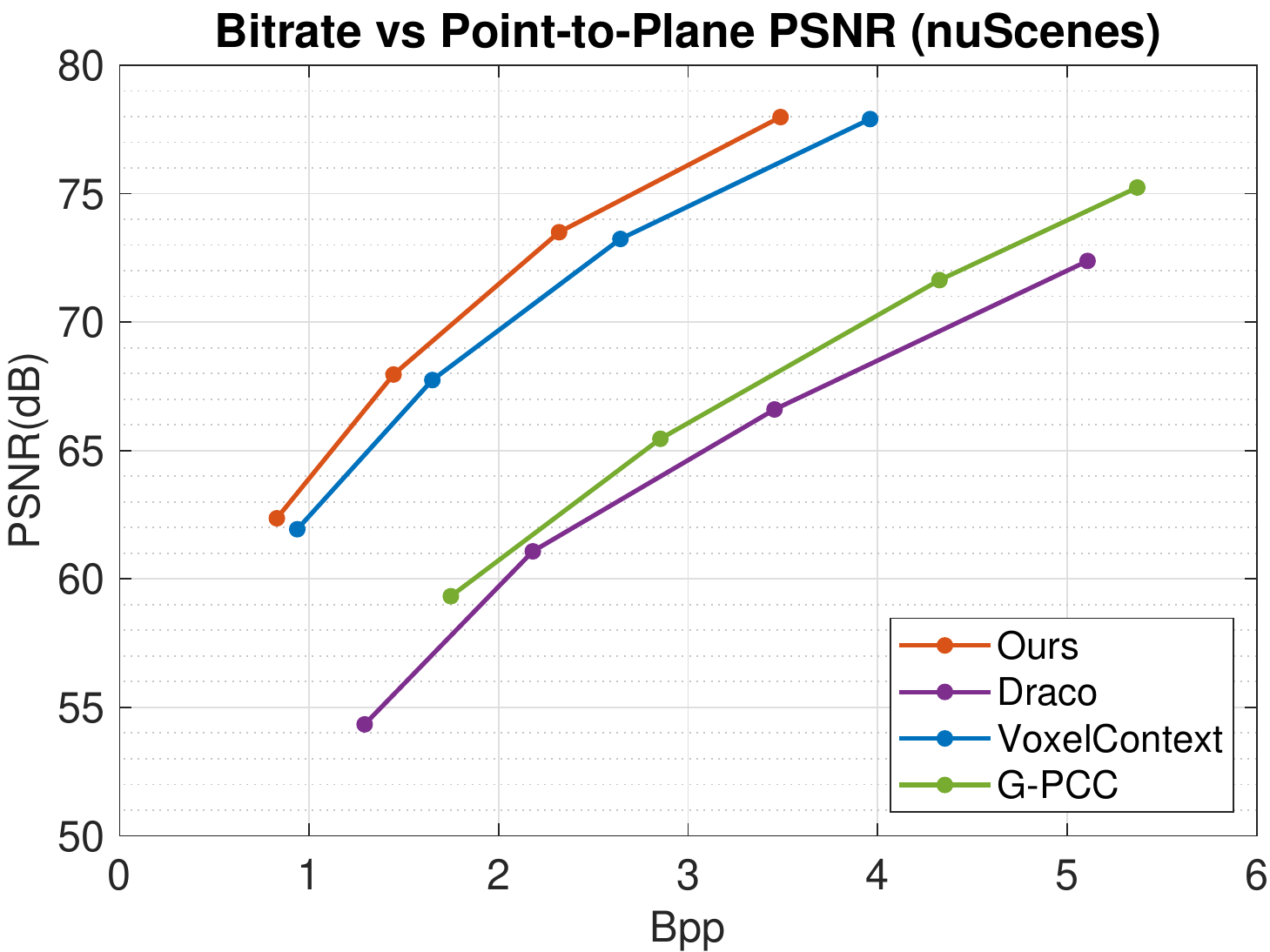}&
\includegraphics[width=0.25\linewidth]{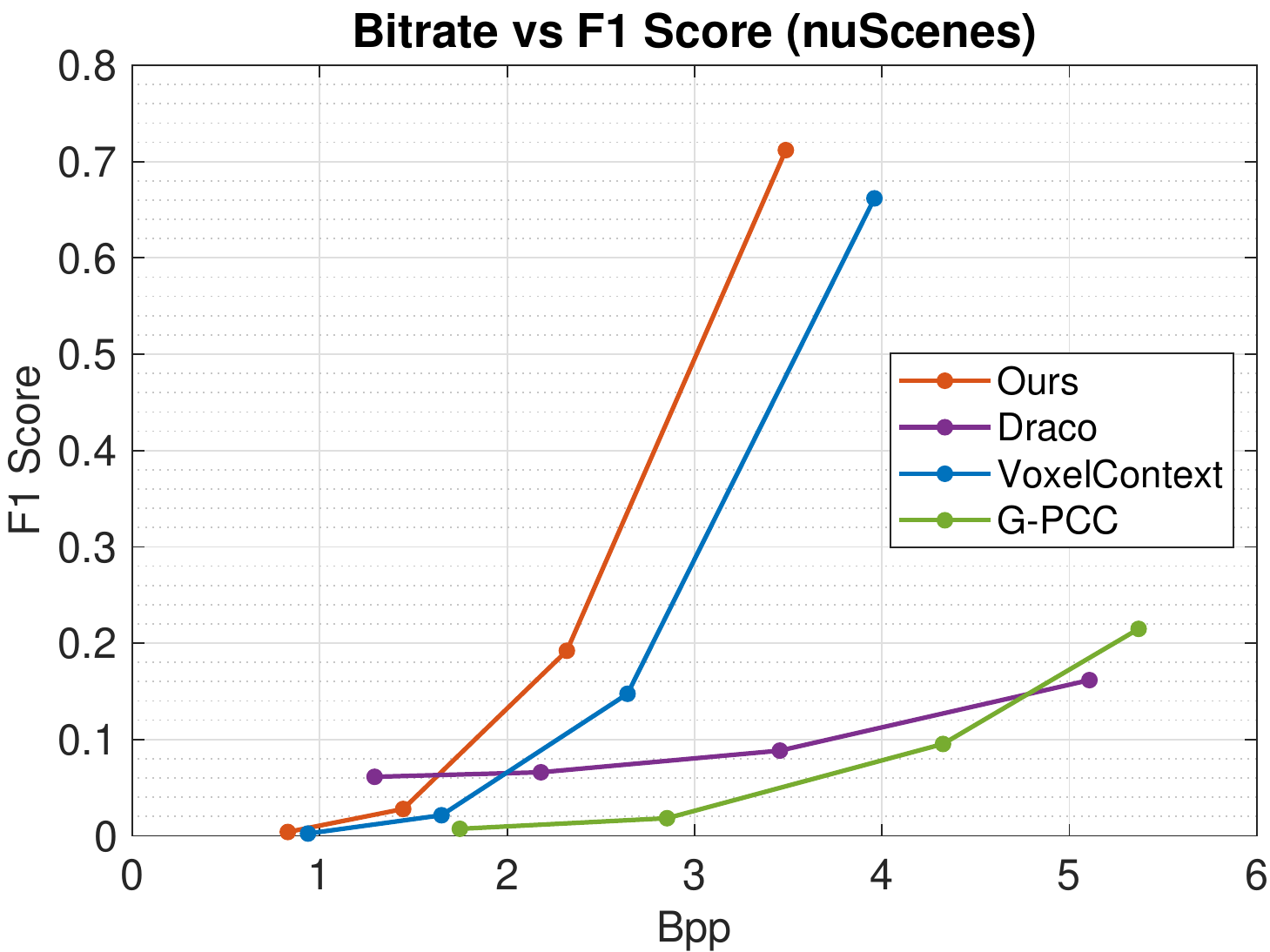}&\\
\end{tabular}
\vspace{-1.5em}
\caption{The quantitative results of our method on the KITTI Odometry dataset (first row) and nuScenes dataset (second raw). From left to right: maximum Chamfer distance ($\downarrow$), point-to-point PSNR ($\uparrow$), point-to-plane PSNR ($\uparrow$), and F1 score ($\uparrow$). Note that both our method and VoxelContext are applied to the refinement module when calculating the evaluation metrics in this figure.}
\label{fig:quantitative}
\end{figure*}

\textbf{Quantitative results.} To quantitatively evaluate our method, we report BPP versus four reconstruction quality metrics. As illustrated in Fig.~\ref{fig:quantitative}, our method saves more bitrate and achieves higher reconstruction quality compared with all the state-of-the-art methods on both 64 channel KITTI Odometry dataset and 32 channel nuScenes dataset.

\begin{table}[ht!]
\centering
\setlength{\tabcolsep}{1.2mm}
\begin{tabular}{@{}l@{\hspace{4mm}}c@{\hspace{4mm}}c@{\hspace{4mm}}c@{\hspace{4mm}}c@{\hspace{4mm}}c@{\hspace{4mm}}c@{\hspace{1mm}}}
\toprule

\multirow{2}{*}{Dataset} & \multirow{2}{*}{Method} &\multicolumn{5}{@{\hspace{-1mm}}c}{BPP$\downarrow$} \\
{} & {} & Level 8 & Level 9 & Level 10 & Level 11 & Level 12\\
\midrule
\multirow{2}{*}{KITTI} & 
VoxelContext ~\cite{que2021voxelcontext} & 0.173 &  0.466 & 1.123 & 2.380 & 4.371 \\ 
& Ours  & \textbf{0.149} & \textbf{0.409} & \textbf{0.999} & \textbf{2.137}  & \textbf{3.878} \\
\bottomrule

\multirow{2}{*}{nuScenes} & VoxelContext~\cite{que2021voxelcontext} & 0.474 & 0.937 & 1.650 & 2.642 & 3.960 \\
& Ours & \textbf{0.424} & \textbf{0.829} & \textbf{1.445} & \textbf{2.319}  & \textbf{3.487} \\

\bottomrule
\end{tabular}
\vspace{+1.0em}
\caption{The quantitative results when compared to VoxelContext without any refinement module. The first row is the result of the KITTI Odometry dataset. The second row is the result of the nuScenes dataset. The reconstructed point clouds of two methods are the same at each level.}
\label{table:voxel}
\end{table}

Since we use the same octree construction strategy as VoxelContext, we have the same reconstruction quality at the same octree level without applying any refinement module. As illustrated in Table~\ref{table:voxel}, our method saves 11\%-16\% bitrate on the KITTI Odometry dataset and 12\%-14\% on the nuScenes dataset compared to VoxelContext. Our method reaches higher reconstruction quality than VoxelContext at the same octree level despite using the same refinement module, clearly demonstrating the benefit of our newly proposed two-step refinement strategy.

\begin{figure*}[t]
\centering
\begin{tabular}{@{}c@{\hspace{1mm}}c@{\hspace{1mm}}c@{\hspace{1mm}}c@{}}
\rotatebox{90}{\small \hspace{4mm} KITTI}
\includegraphics[width=0.235\linewidth]{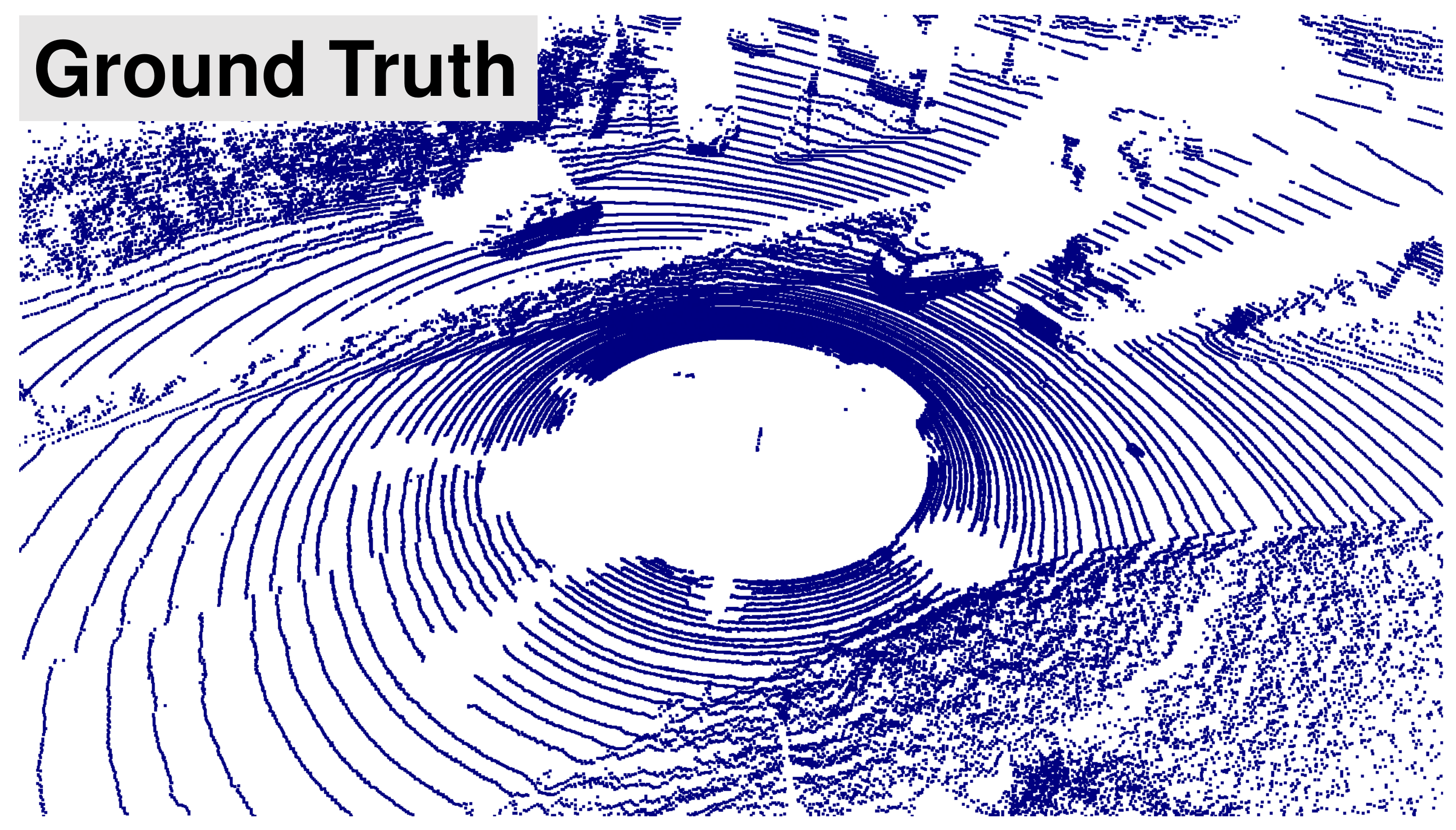}&
\includegraphics[width=0.235\linewidth]{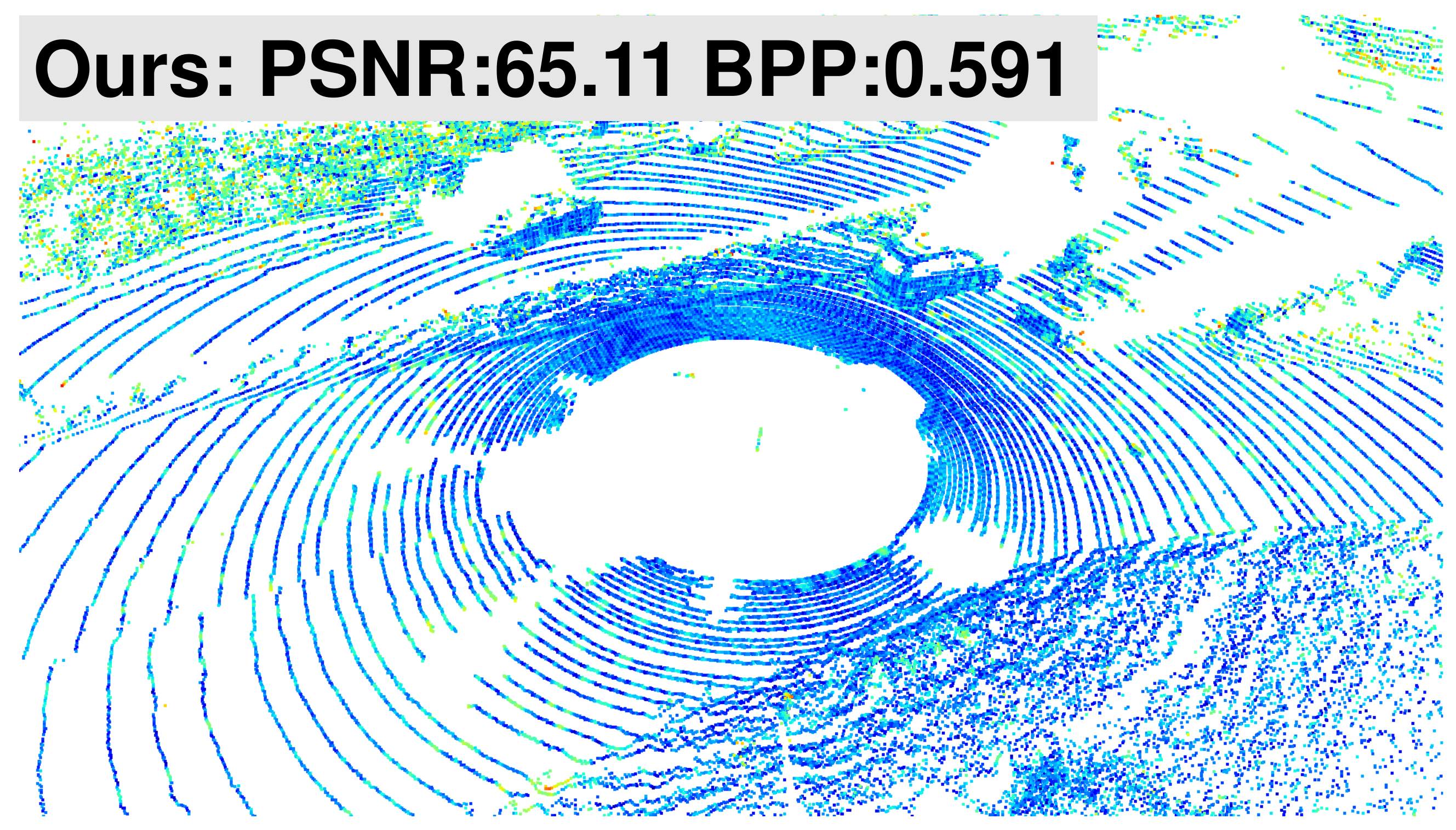}&
\includegraphics[width=0.235\linewidth]{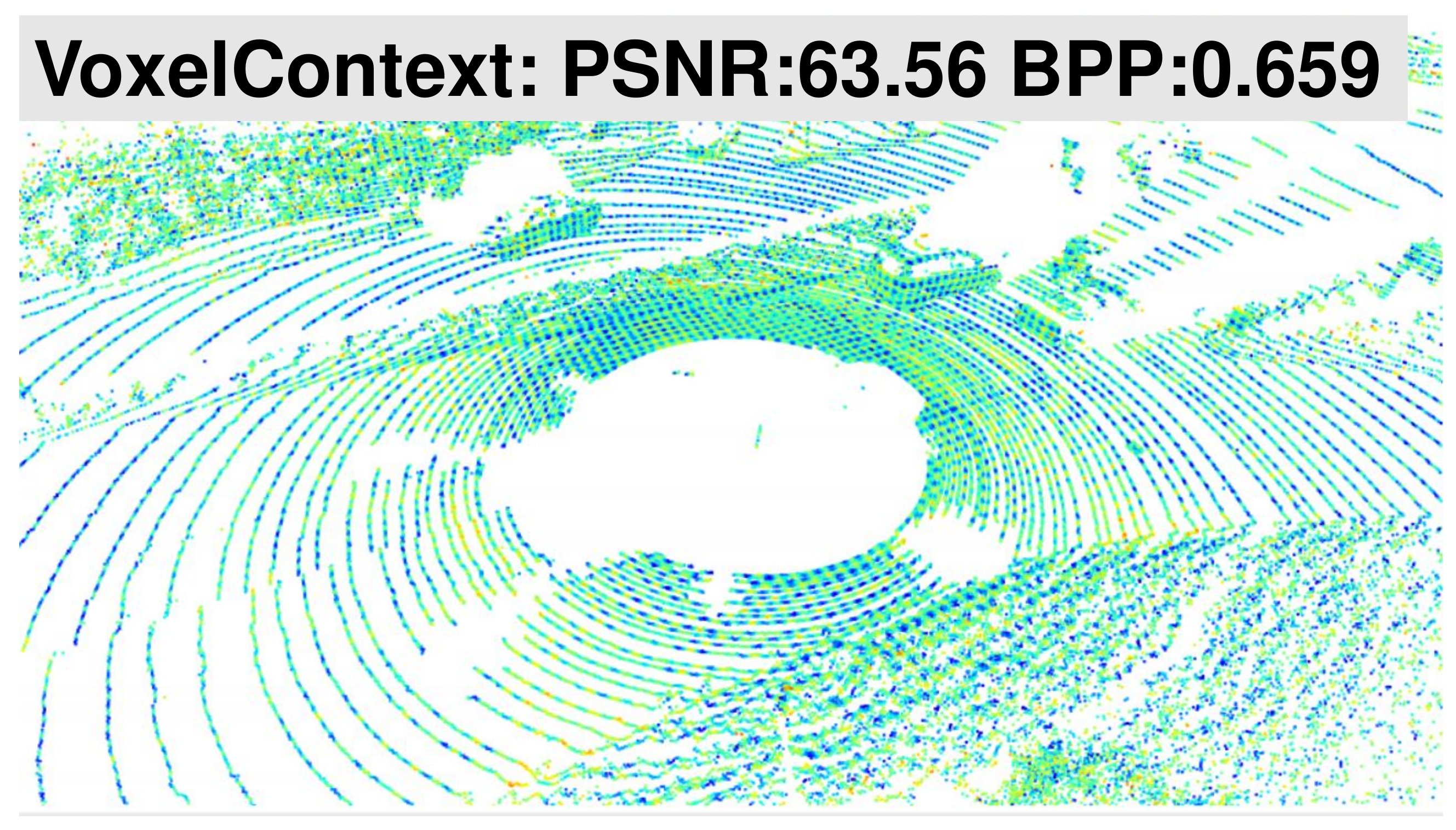}&
\includegraphics[width=0.235\linewidth]{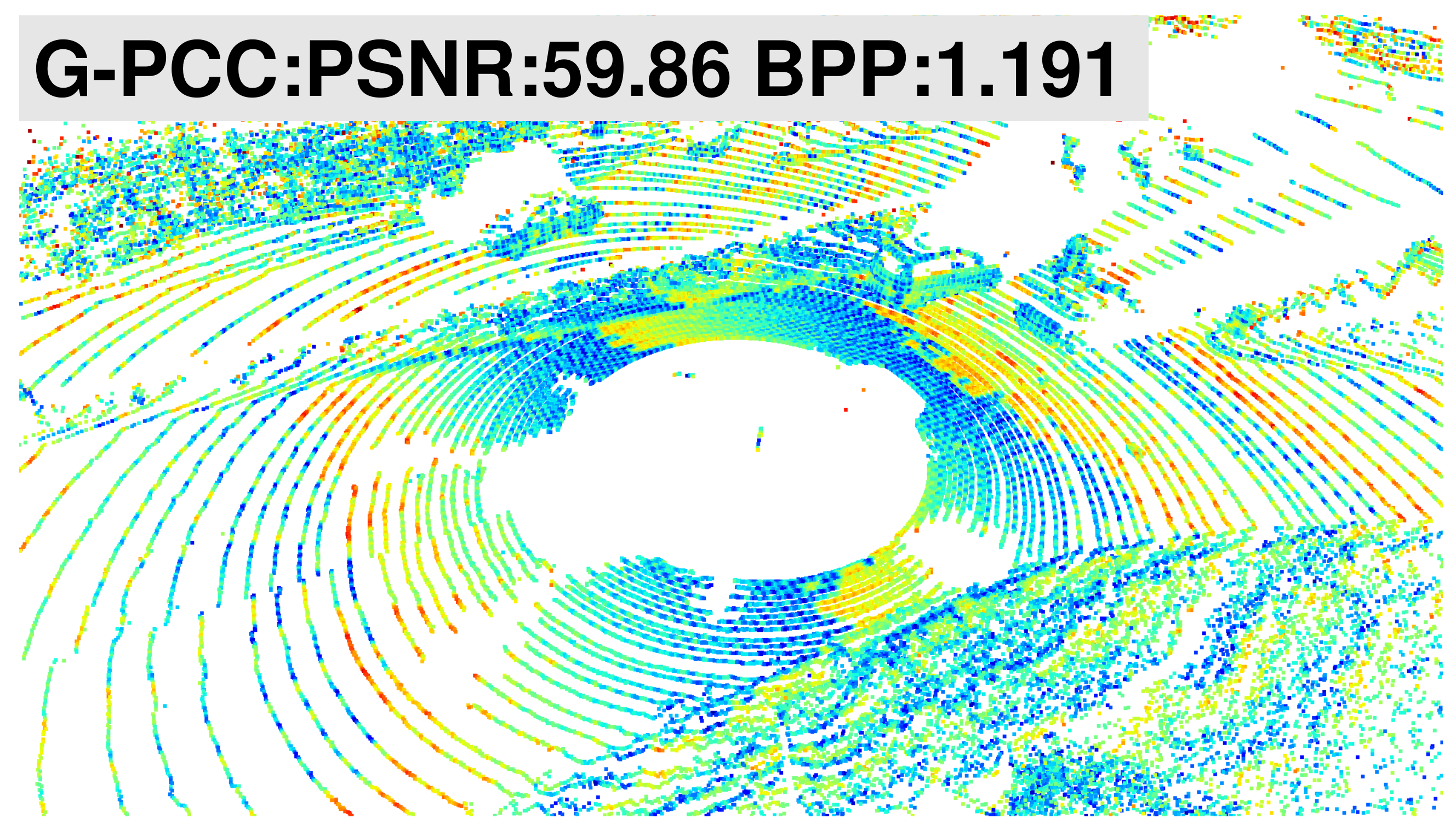}\\

\rotatebox{90}{\small nuScenes}
\includegraphics[width=0.235\linewidth]{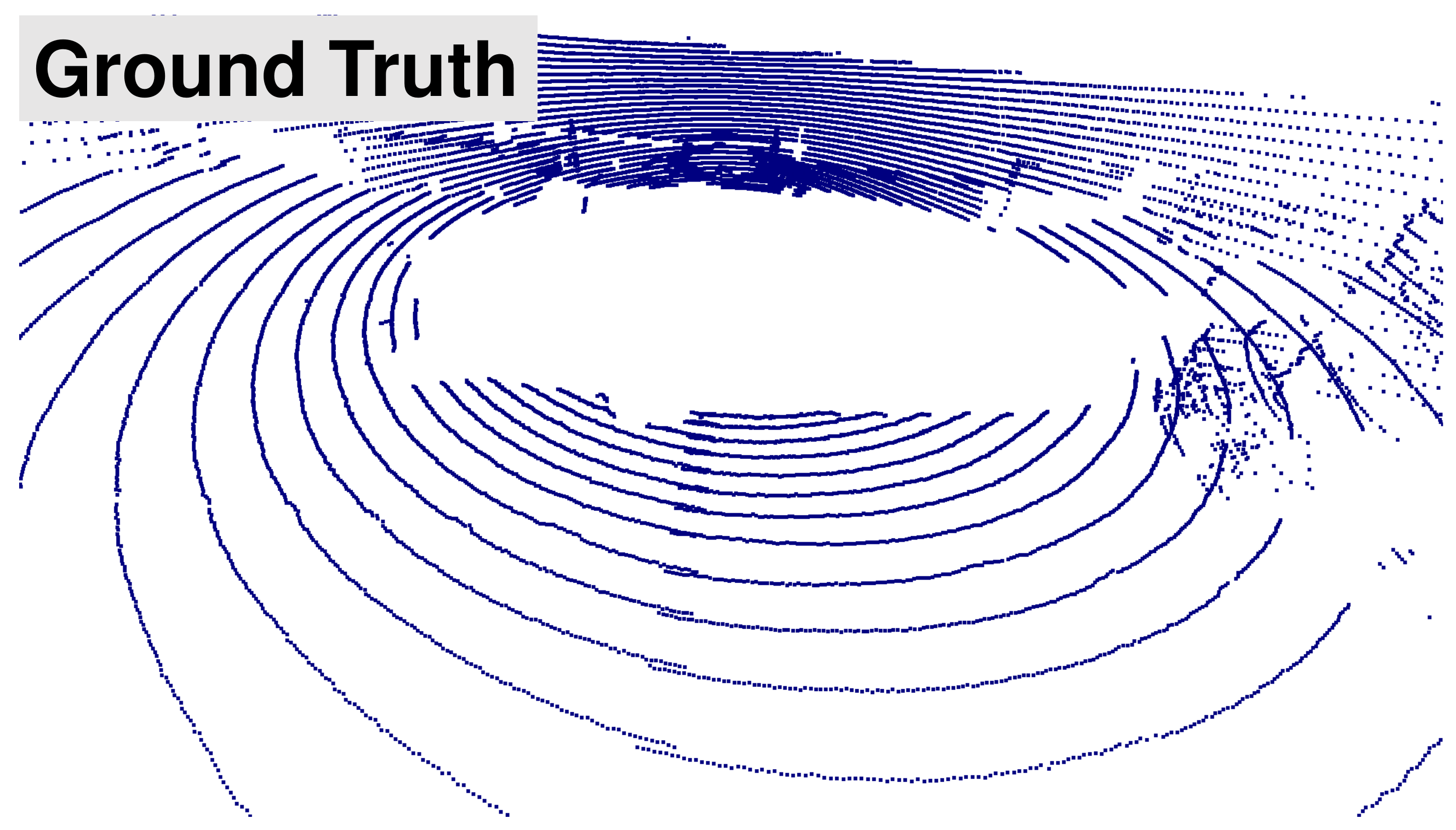}&
\includegraphics[width=0.235\linewidth]{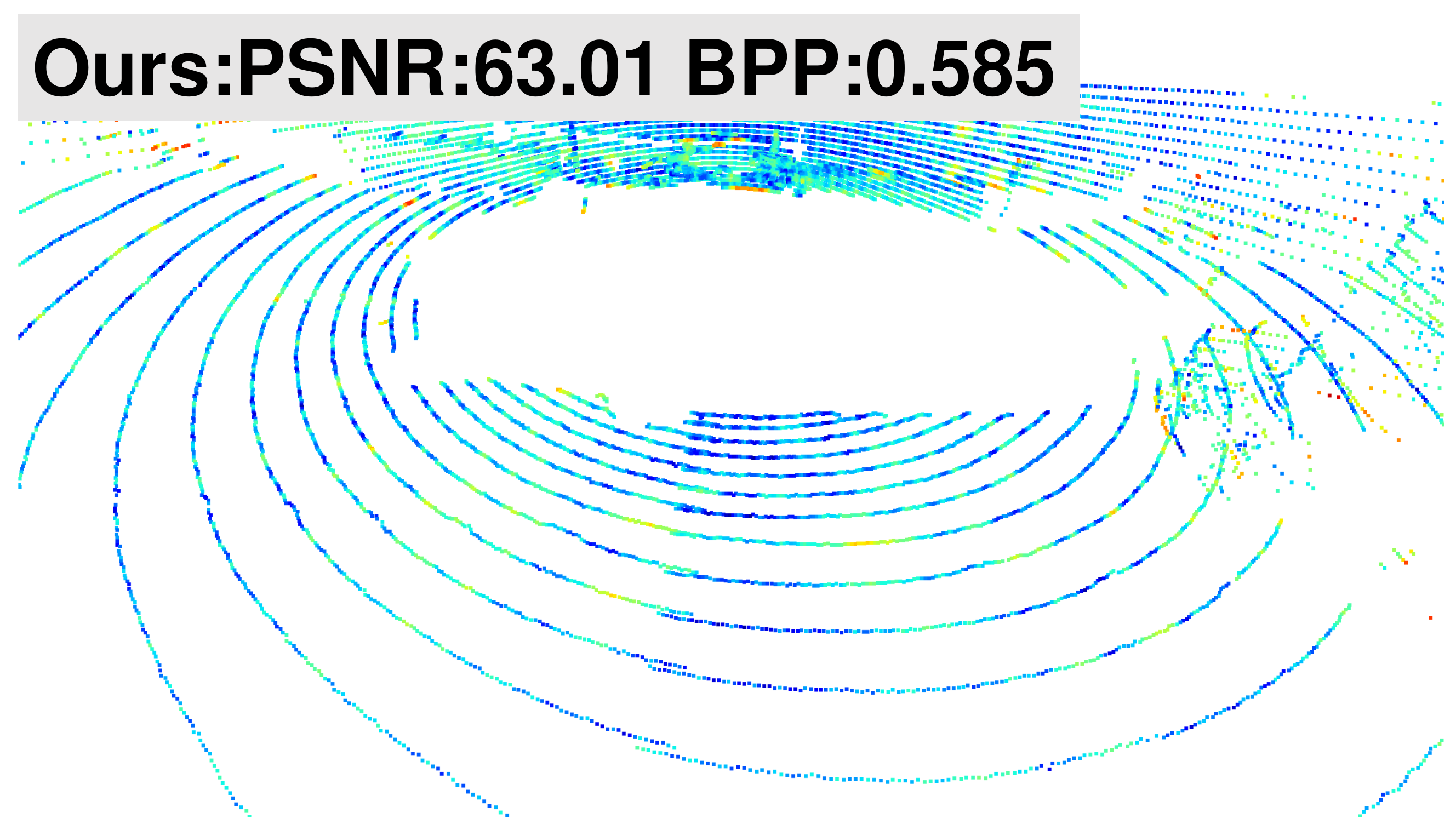}&
\includegraphics[width=0.235\linewidth]{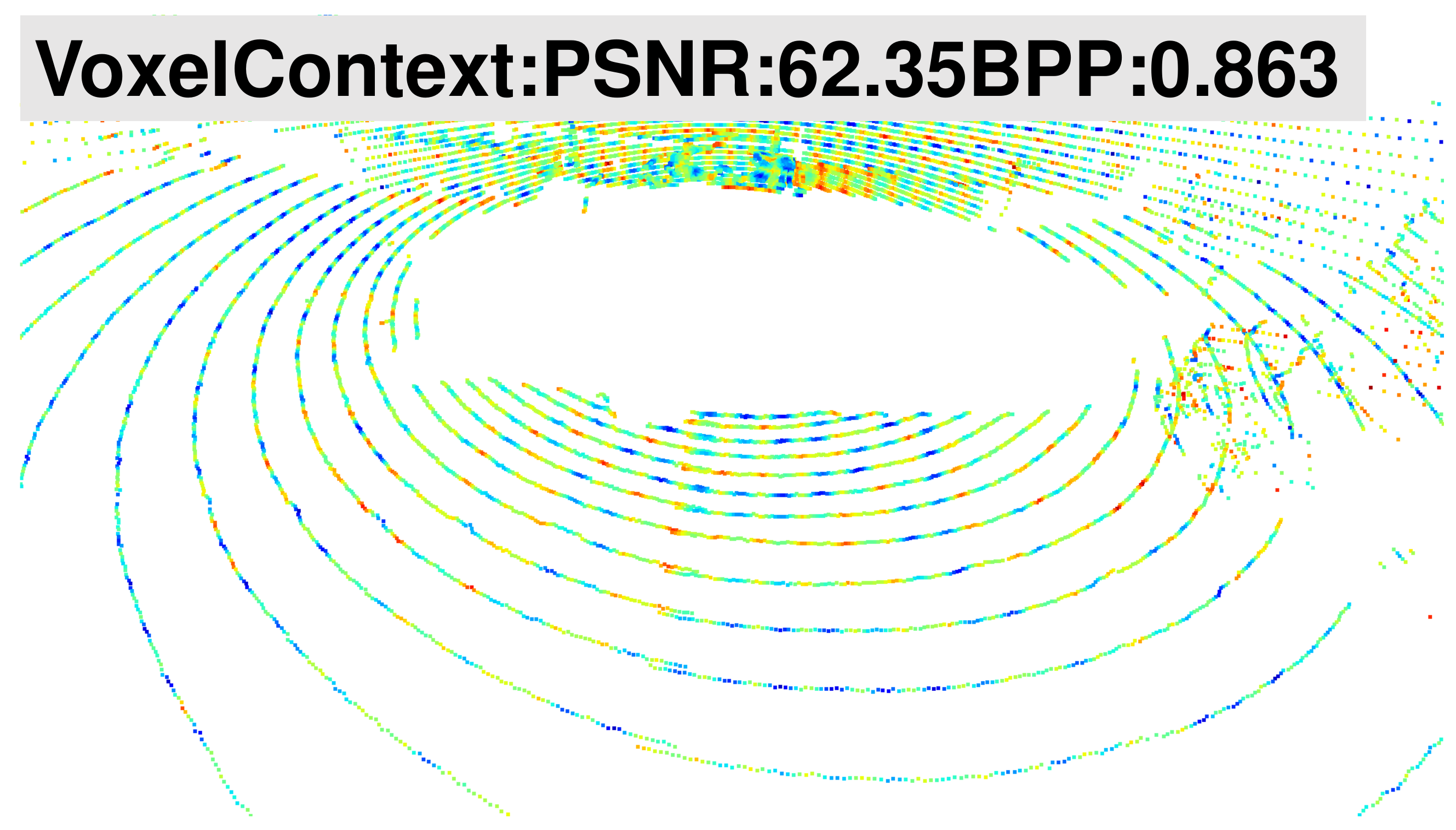}&
\includegraphics[width=0.235\linewidth]{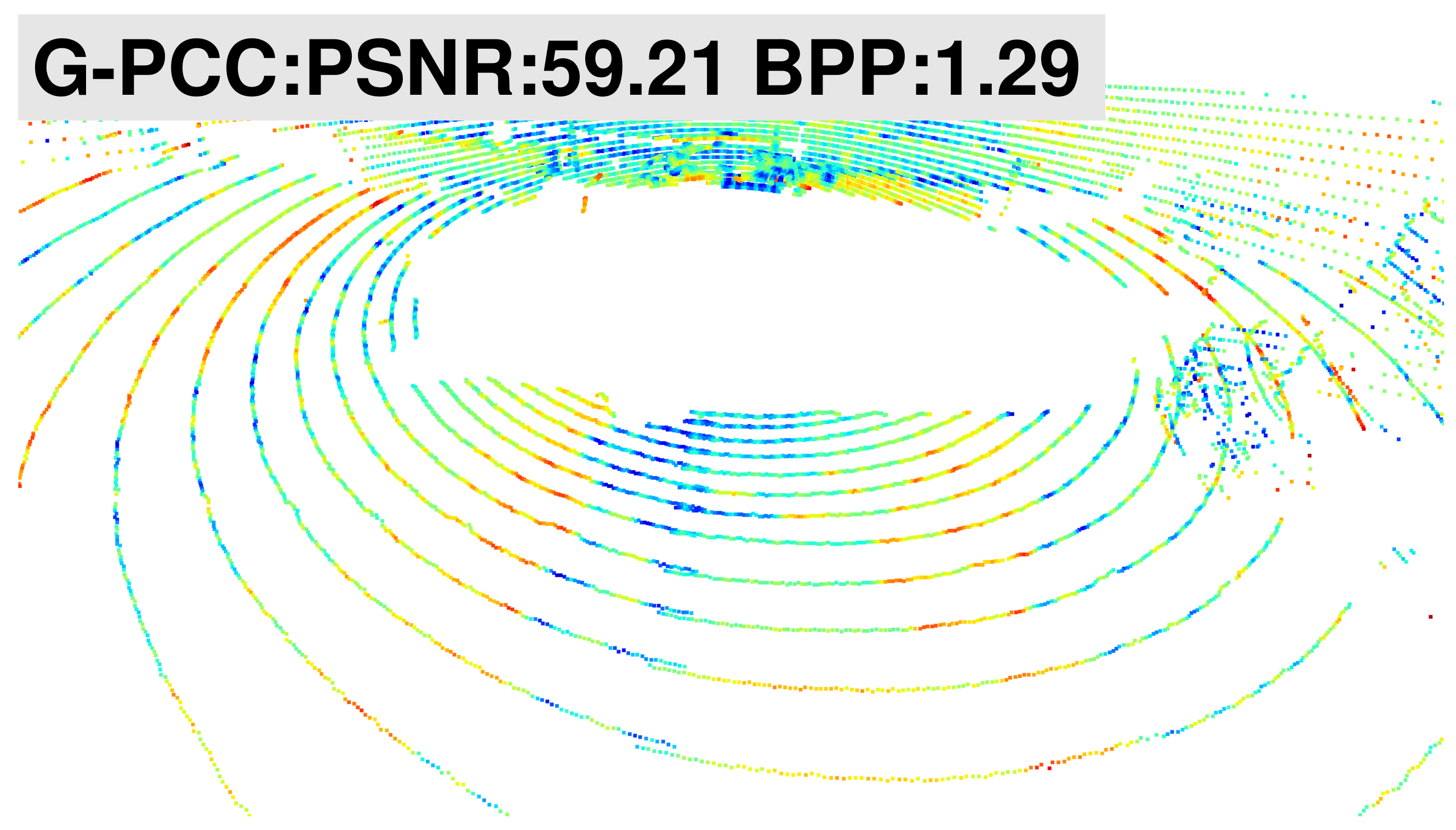}\\
\end{tabular}
\includegraphics[width=1.0\linewidth]{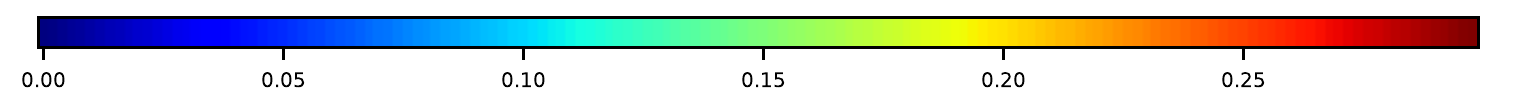}\\

\vspace{-1.0em}
\caption{The qualitative results of our method compared with VoxelContext and G-PCC on the KITTI Odometry dataset (first row) and the nuScenes dataset (second row). The color bar indicates the error in meters between the original and reconstructed point cloud. Our reconstructed point cloud has a lower error at a lower bitrate compared to all baselines.}
\label{fig:qualitative}
\end{figure*}

\begin{figure}
\centering
\begin{tabular}{cc}

\includegraphics[width=0.35\linewidth]{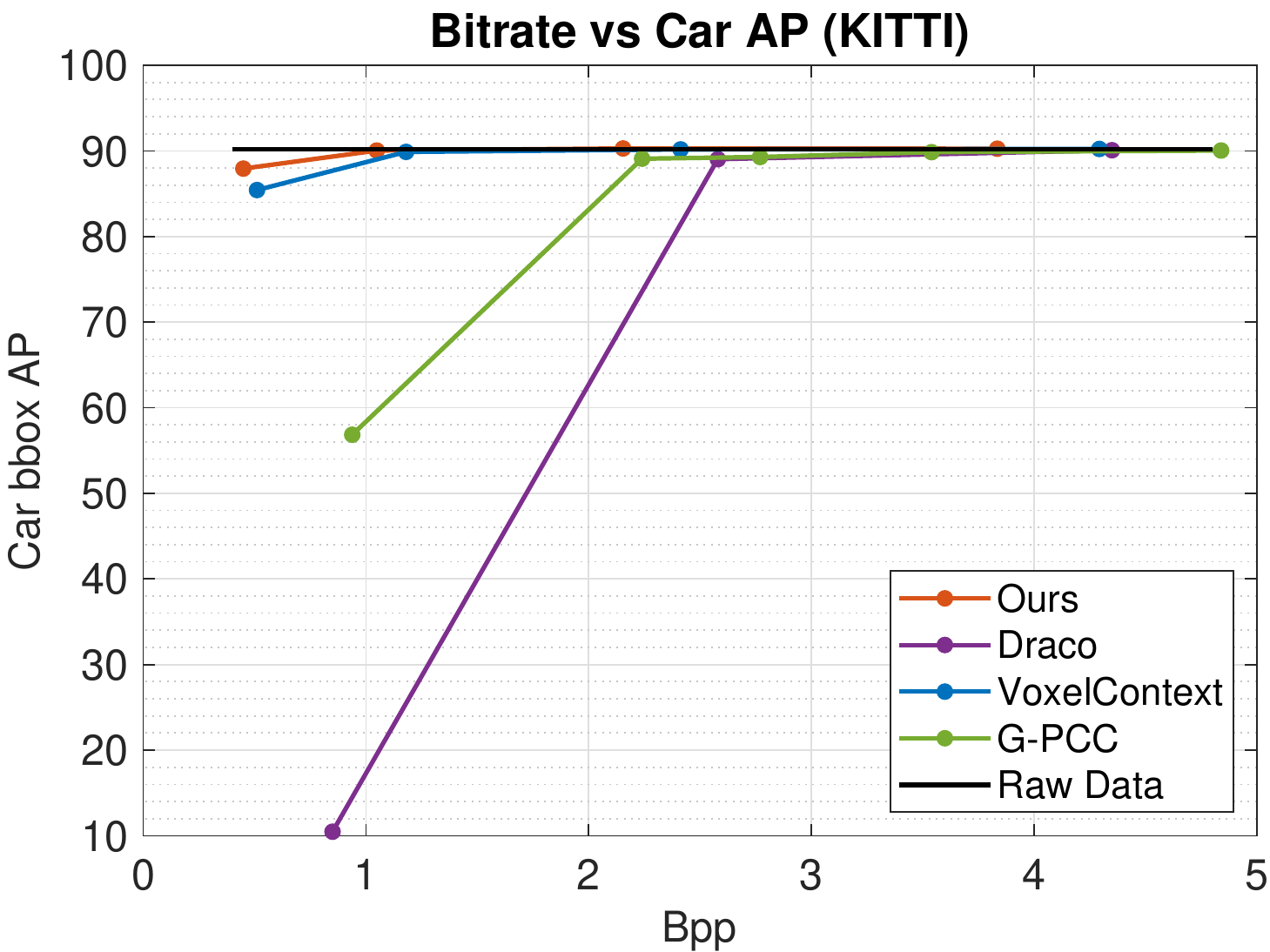}&
\includegraphics[width=0.35\linewidth]{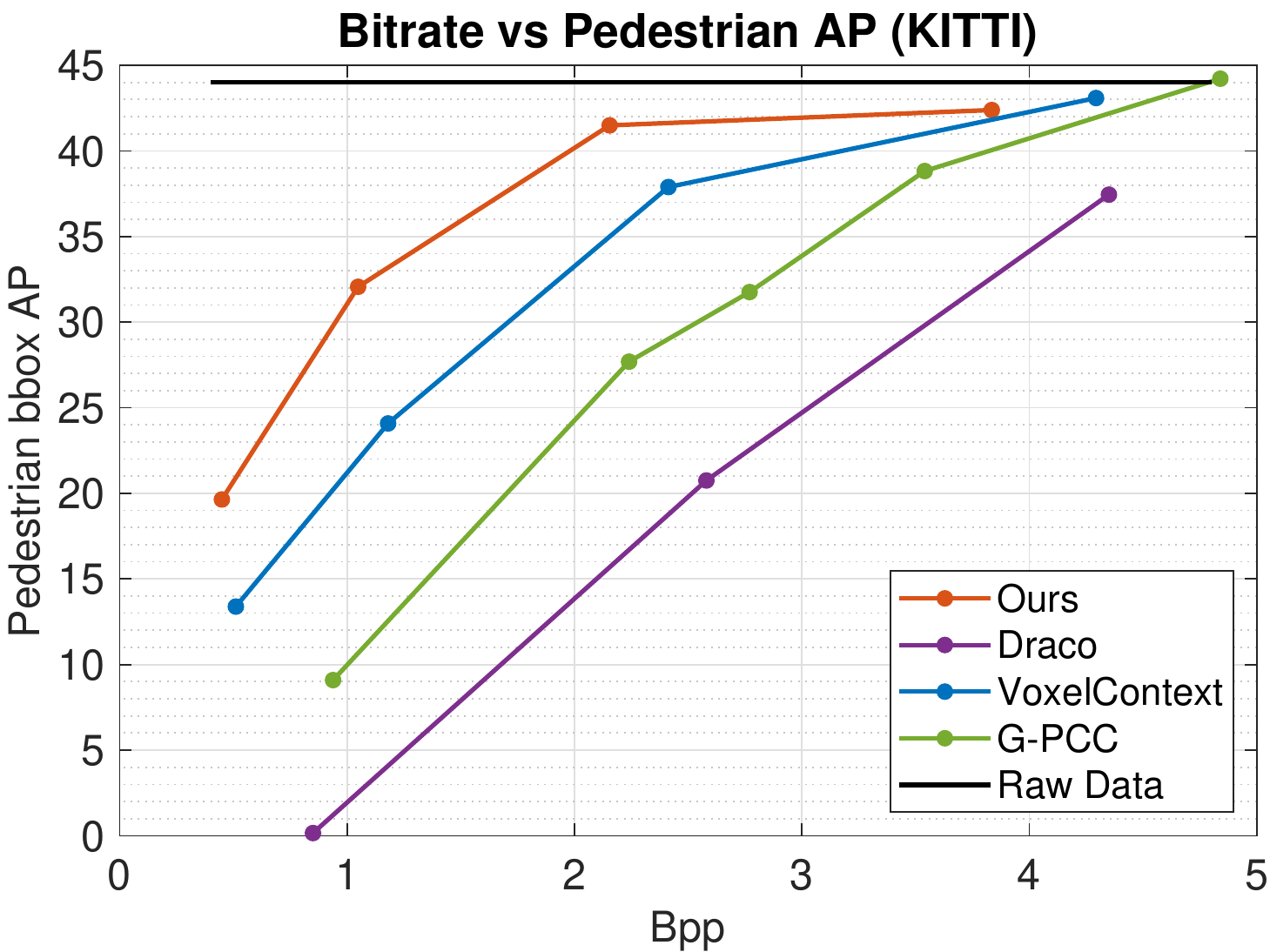}\\
\end{tabular}
\vspace{-1.0em}
\caption{The qualitative results of downstream tasks on the KITTI detection dataset.}
\label{fig:downstream}
\end{figure}

\begin{figure*}[t]
\centering
\begin{tabular}{@{}c@{\hspace{0.2mm}}c@{\hspace{0.2mm}}c@{\hspace{0.2mm}}c@{\hspace{0.2mm}}c@{}}
\hspace{-2mm}
\rotatebox{90}{\tiny Apollo-DaoxiangLake}
\includegraphics[width=0.25\linewidth]{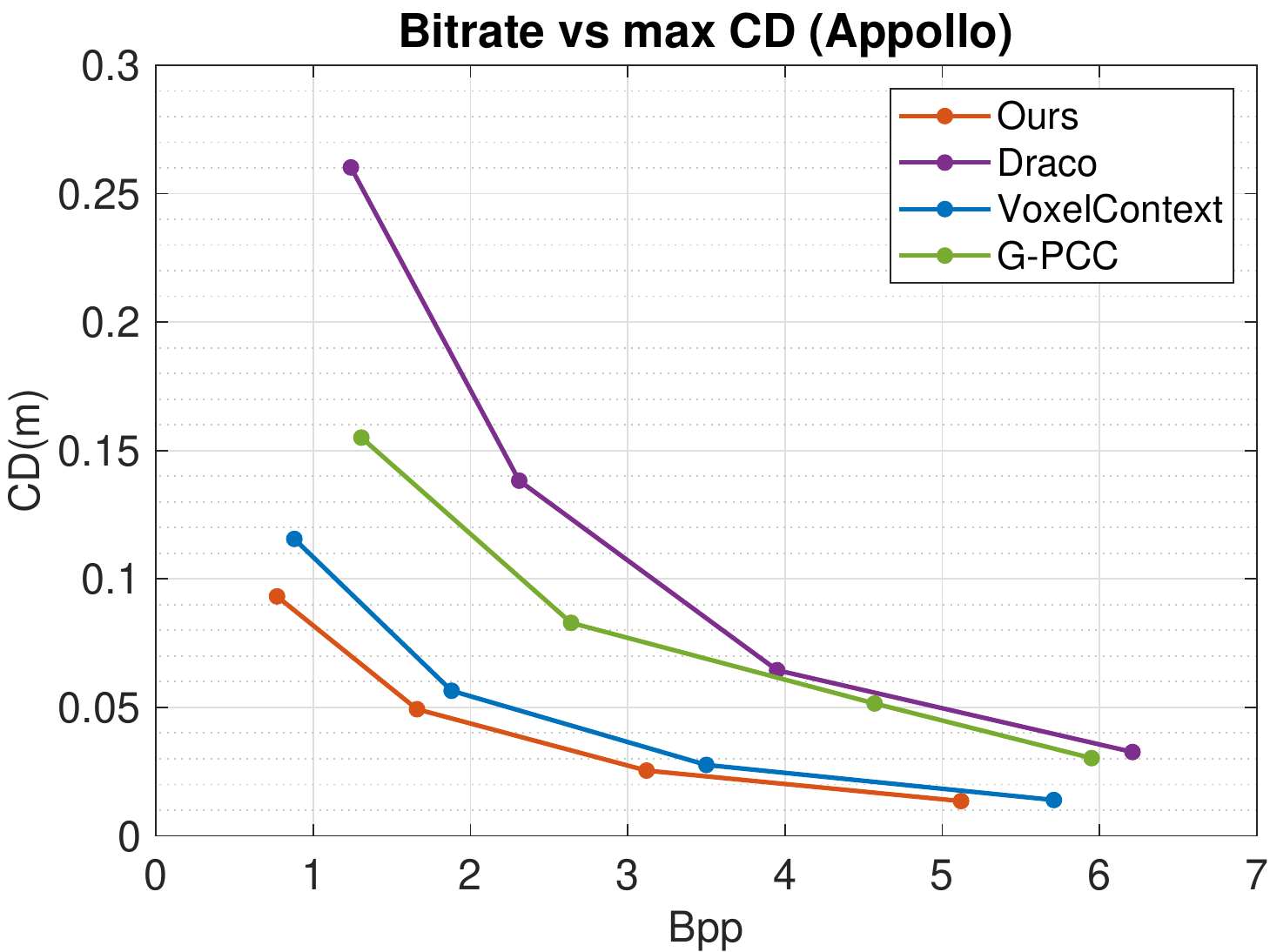}&
\includegraphics[width=0.25\linewidth]{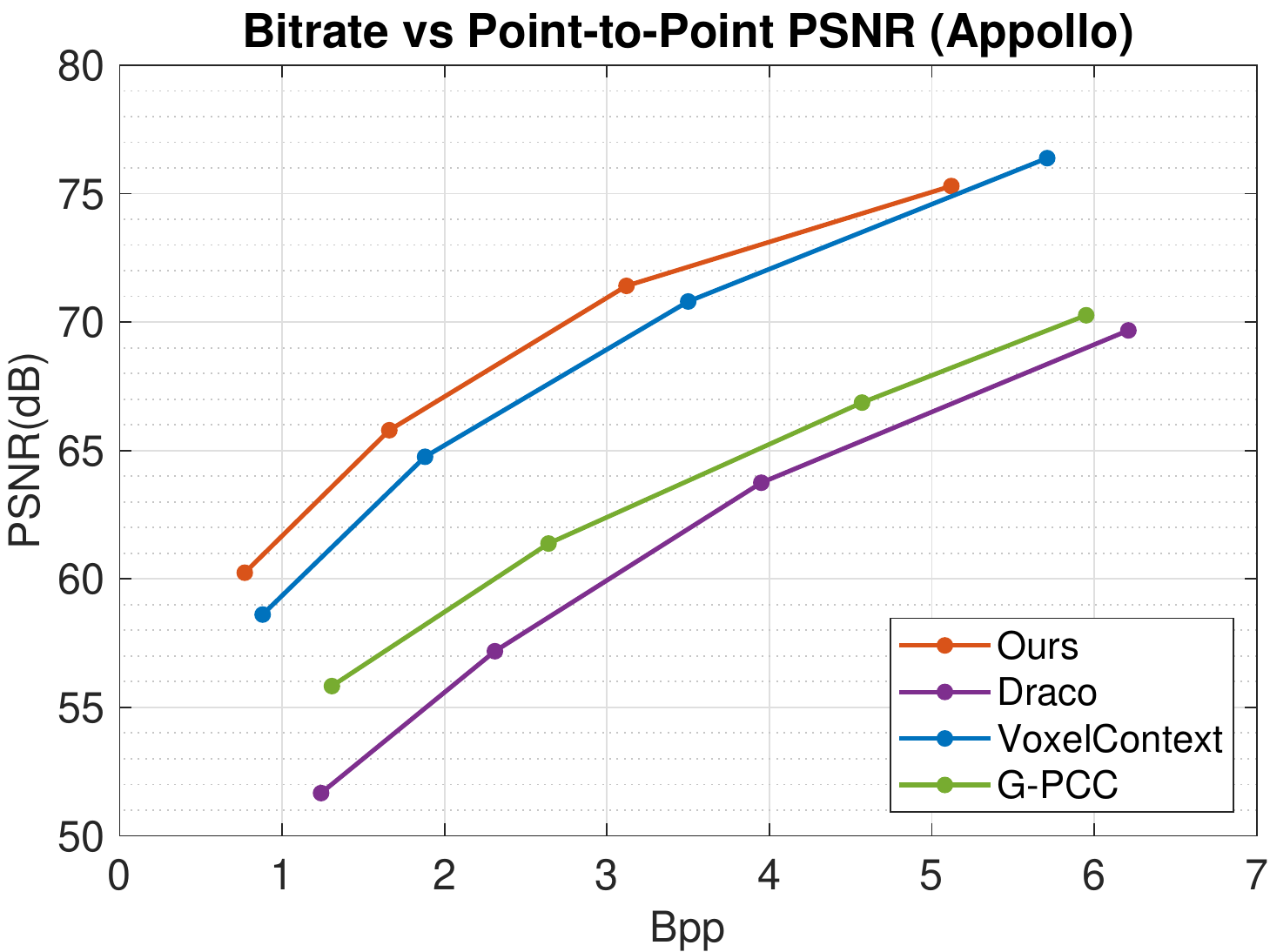}&
\includegraphics[width=0.25\linewidth]{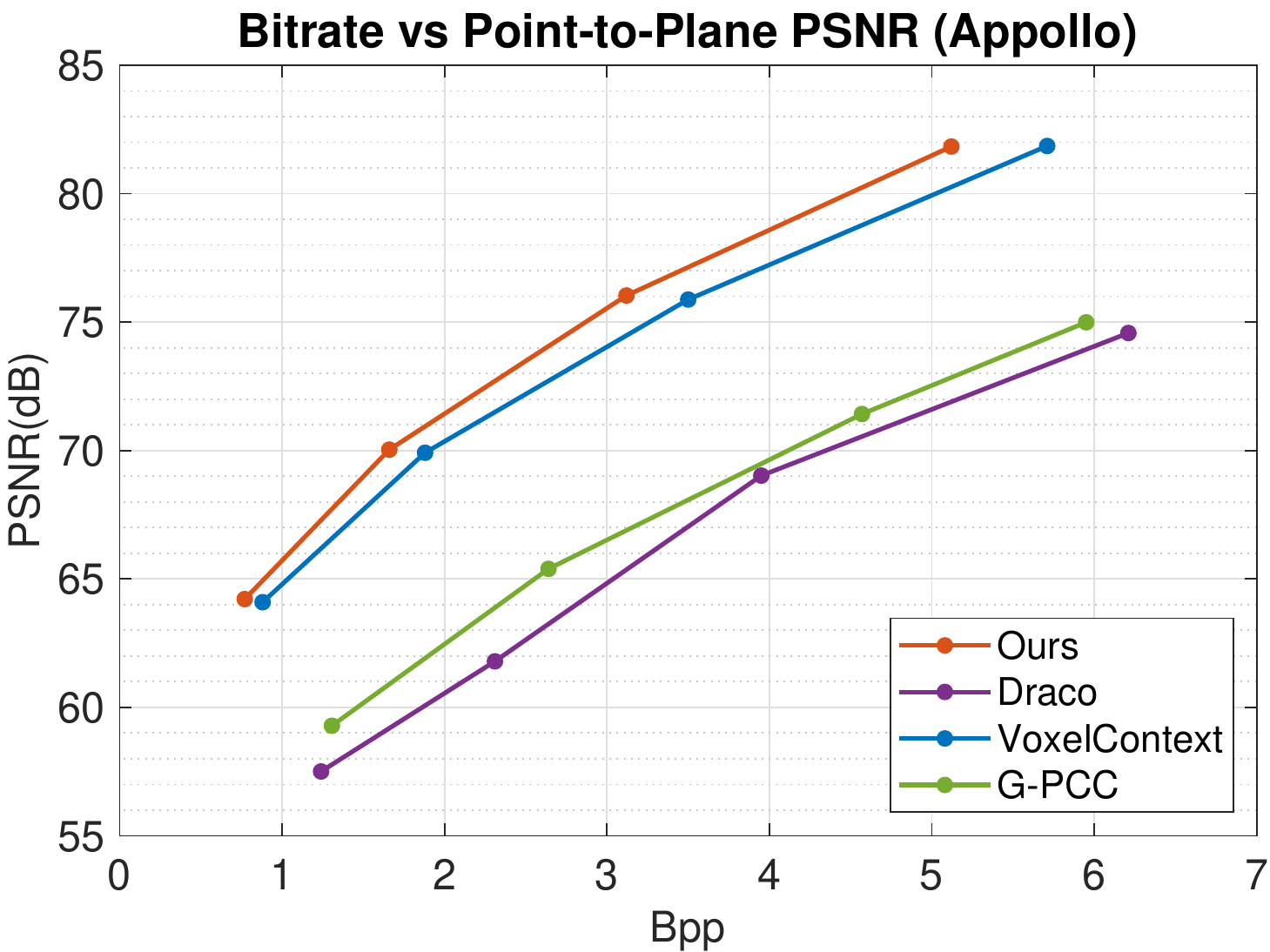}&
\includegraphics[width=0.25\linewidth]{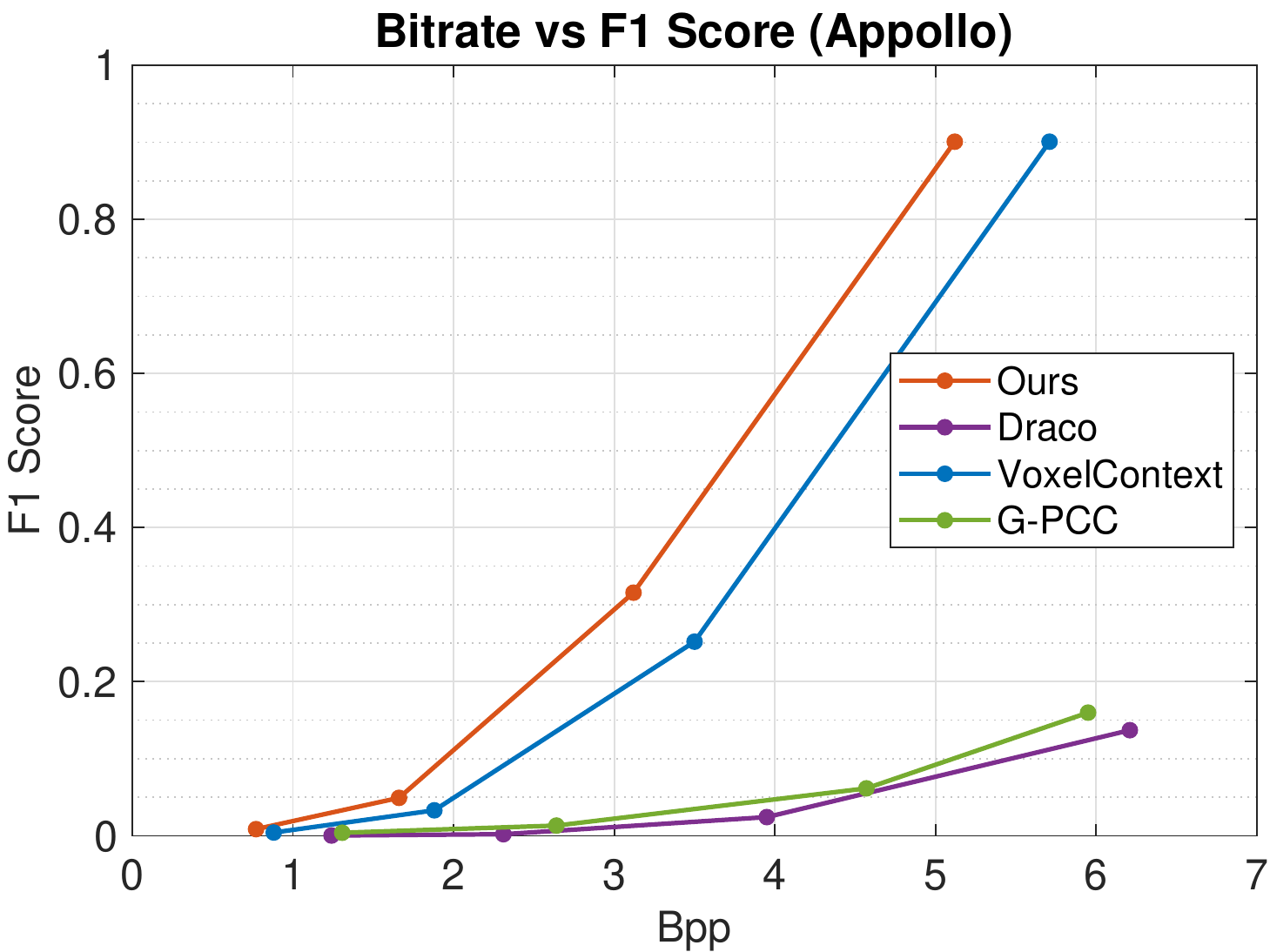}\\
\end{tabular}
\vspace{-1.5em}
\caption{The cross-dataset quantitative results of our method on the Apollo-DaoxiangLake dataset. The result shows that our model has better generalization ability than baselines.}
\label{fig:crossdata}
\end{figure*}

\textbf{Qualitative results.}
We evaluate our method qualitatively by comparing the reconstruction error with state-of-the-art methods on two datasets. As illustrated in Fig.~\ref{fig:qualitative}, the error between the original point cloud and our reconstructed point cloud is much smaller at a even lower bit rate as compared to all three baseline methods.

\textbf{Downstream tasks.}
The performance on downstream tasks of the reconstructed point cloud is also a crucial evaluation metric for point cloud compression. We evaluate the 3D object detection method proposed in~\cite{lang2019pointpillars} on reconstructed point cloud data of different compression methods. We report the bitrate versus average precision (AP) at 0.5 IOU as threshold for pedestrians and 0.7 IOU for cars.
As illustrated in Fig.~\ref{fig:downstream}, our method outperforms all three baselines on different categories.

\textbf{Cross-dataset results.}
To better show our model's generalization ability, we evaluate the cross-dataset performance on the Apollo-DaoxiangLake dataset with our model trained on the KITTI Odometry dataset. The Apollo-DaoxiangLake dataset captures point clouds in different driving scenes with the same type of LiDAR as the KITTI Odometry dataset. As shown in Fig.~\ref{fig:crossdata}, the evaluation results show our models still have competitive compression performance over the baselines with better reconstruction quality. In terms of bitrates, our method saves 11.2-13.7\% compared to VoxelContext, 56.91-137.35\% to Draco, and 68.83-125.08\% to G-PCC, respectively.

\subsection{Ablation Study}
We perform an ablation study on our deep entropy model to demonstrate the effectiveness of different contextual features. We ablate over the contextual features by training the model without exploiting features from siblings' children, ancestors, or surfaces. The models for ablation studies are trained on the KITTI Odometry dataset with the same training configurations. We evaluate the models at a level ranging from 8 to 12. From the quantitative results shown in Table~\ref{table:Ablation}, we observe that the bitrate increases when not incorporating any dependencies into the entropy model. Moreover, the result also demonstrates that the sibling features are crucial to the compression performance of our entropy model. 

\begin{table*}[ht!]
\small
\centering
\setlength{\tabcolsep}{3mm}

\begin{tabular}{@{}l@{\hspace{1mm}}c@{\hspace{1mm}}c@{\hspace{1mm}}c@{\hspace{1mm}}c@{\hspace{1mm}}c@{}}
\toprule

\multirow{2}{*}{Method}&\multicolumn{5}{@{\hspace{-1mm}}c}{BPP$\downarrow$} \\
& Level 8 & Level 9 & Level 10 & Level 11 & Level 12\\
\midrule
Ours w/o Sibling & \footnotesize 0.161 \tiny(+8.1$\%$) & \footnotesize 0.439 \tiny(+7.3\%) & \footnotesize 1.071 \tiny(+7.2\%) & \footnotesize 2.288 \tiny(+7.1\%) & \footnotesize 4.133 \tiny(+6.6\%) \\

Ours w/o Ancestor & \footnotesize 0.150 \tiny(+0.6$\%$) & \footnotesize 0.414 \tiny(+1.2\%) & \footnotesize 1.016 \tiny(+1.7\%) & \footnotesize 2.181 \tiny(+2.1\%) & \footnotesize 3.956 \tiny(+2.0\%) \\

Ours w/o Surface & \footnotesize 0.151 \tiny(+1.3$\%$) & \footnotesize 0.415 \tiny(+1.5\%) & \footnotesize 1.019 \tiny(+2.0\%) & \footnotesize 2.180 \tiny(+2.0\%) & \footnotesize 3.953 \tiny(+1.9\%) \\
\midrule
\textbf{Ours} & \textbf{0.149} & \textbf{0.409} & \textbf{0.999} & \textbf{2.137}  & \textbf{3.878} \\
\bottomrule
\end{tabular}
\vspace{+1.0em}
\caption{Ablation study of our entropy model on KITTI~\cite{Geiger2012CVPR} without using context from siblings' children, ancestors, or surface priors.}
\label{table:Ablation}
\end{table*}

\section{Conclusion}
We have presented a novel octree-based point cloud compression method. Our method uses the hierarchical contexts of octree structures and surface priors to reduce the information redundancies in the bitstream. On the decoding side, our method utilizes a two-step heuristic strategy for reconstructing point cloud with better quality. We have evaluated our method against three state-of-the-art baselines on two datasets. The result demonstrates that our method outperforms previous methods on compression performance, reconstruction quality, and downstream tasks. We hope our work can inspire researchers to further reduce the compression rate and improve the reconstruction quality in future work.

\clearpage
%
%
\bibliographystyle{splncs04}
\bibliography{egbib}
\end{document}